\documentclass[11pt]{article}

\usepackage[preprint]{acl}   

\usepackage{times}
\usepackage{latexsym}
\usepackage[T1]{fontenc}
\usepackage[utf8]{inputenc}
\usepackage{microtype}
\usepackage{inconsolata}

\usepackage{amsmath,amsfonts,amssymb}
\usepackage{mathtools}
\usepackage{graphicx}
\usepackage{booktabs}
\usepackage{tabularx}
\usepackage{multirow}
\usepackage{makecell}
\usepackage{placeins}
\usepackage{enumitem}
\usepackage{algorithm}
\usepackage{algpseudocode}
\usepackage{xcolor}
\usepackage{listings}
\usepackage{tikz}
\usetikzlibrary{positioning,arrows.meta,calc,shapes.geometric,fit,shadows,decorations.pathreplacing}
\hypersetup{hidelinks}

\usepackage{comment}

\lstdefinestyle{prompttemplate}{
  basicstyle=\ttfamily\footnotesize,
  breaklines=true,
  breakatwhitespace=false,
  columns=fullflexible,
  keepspaces=true,
  frame=single,
  framerule=0pt,
  rulesep=0pt,
  backgroundcolor=\color{black!3},
  numbers=none,
  xleftmargin=0.5em,
  xrightmargin=0.5em,
  framexleftmargin=6pt,
  framexrightmargin=6pt,
  framesep=6pt,
  aboveskip=0.5\baselineskip,
  belowskip=0.5\baselineskip
}

\newif\ifanon
\anonfalse
\ifanon
  \newcommand{\sysname}{Sentinel}
  \newcommand{\fmname}{Forecast-FM}
  \newcommand{\benchname}{Decision-Bench}
  \newcommand{\corpusname}{Exposure-Corpus}
\else
  \newcommand{\sysname}{DeXposure-Claw}
  \newcommand{\fmname}{DeXposure-FM}
  \newcommand{\benchname}{DeXposure-Bench}
  \newcommand{\corpusname}{DeXposure}
\fi

\title{\sysname{}: An Agentic System for DeFi Risk Supervision}

\author{Aijie Shu\\
University of Edinburgh\\
\texttt{v1ashu@ed.ac.uk}
\And
Bowei Chen\\
University of Glasgow\\
\texttt{bowei.chen@glasgow.ac.uk}
\And
Wenbin Wu\\
University of Cambridge\\
\texttt{w.wu@jbs.cam.ac.uk}
\AND
Cathy Yi-Hsuan Chen\\
University of Glasgow\\
\texttt{CathyYi-Hsuan.Chen@glasgow.ac.uk}
\And
Fengxiang He\\
University of Edinburgh\\
\texttt{fhe@ed.ac.uk}}

\begin{document}
\maketitle

\begin{abstract}
Decentralized finance exposes supervisors to fast-moving, networked credit risks. General-purpose LLM agents fit this setting poorly: they over-read weak evidence and recommend high-stakes interventions, while existing evaluations offer no regulator-aligned way to measure the resulting false alarms. We introduce \sysname{}, a forecast-grounded agentic supervision system that routes LLM decisions through structured evidence: (1) \fmname{}, a graph time-series foundation model, forecasts future exposure networks; (2) deterministic monitors and stress scenarios then turn those forecasts into typed alerts, attribution signals, and scenario evidence; and (3) data-health and confidence gates constrain escalation before \sysname{} emits auditable supervisory tickets with rationales. We further develop \benchname{}, a six-axis evaluation harness, whose decision axis scores tickets against a regulator-aligned absolute-loss ground truth and an explicit false-intervention rate. Experiments on five years of weekly real data fully support our system. Code is at \url{https://github.com/EVIEHub/DeXposure-Claw}.
\end{abstract}

\section{Introduction}
\label{sec:intro}

Decentralized finance (DeFi) creates fast-moving networks of token-mediated credit exposure among lending protocols, decentralized exchanges, stablecoins, bridges, and yield aggregators. Recent crises such as Terra/Luna, FTX, and SVB/USDC show that shocks can propagate across this network before supervisors can manually inspect raw on-chain data~\citep{VidalTomasEtAl2023FTX}. This motivates decision-support systems that can forecast emerging exposure risk, identify affected protocols, and recommend supervisory responses.

A straightforward approach is to build an LLM agent that reasons directly over on-chain measurements. We argue that this is unsafe for high-stakes supervision: general-purpose LLM agents may produce plausible rationales while over-reading incomplete, stale, or weak evidence, thereby triggering unnecessary high-severity interventions. Existing systemic-risk evaluations also often rank protocols by fractional exposure changes~\citep{BERTOMEU2024105321,gonon2025computingsystemicriskmeasures,li2025hgmae}, which can overemphasize small protocols and fail to reflect regulator-relevant loss priorities.

We introduce \emph{\sysname{}}, a forecast-grounded agentic system for DeFi risk monitoring and decision recommendation. Rather than asking an LLM to reason over raw transactions, the system routes decisions through structured forecasted evidence. The graph time-series foundation model \fmname{}~\citep{ShuEtAl2026DeXposureFM} forecasts future exposure networks; deterministic monitors and stress scenarios convert these forecasts into alerts, attribution information, scenario losses, and uncertainty estimates; an LLM then emits ranked supervisory tickets with targets, severities, and evidence-linked rationales. Data-health and confidence gates restrict intervention-level recommendations, and every ticket is released with its evidence bundle, rationale, and gate states for auditability.

We evaluate the system with \benchname{}, a six-axis harness covering forecast quality, warning behavior, uncertainty calibration, stress-scenario fidelity, ticket quality, and robustness. Its decision axis uses a regulator-aligned absolute-loss ground truth, allowing false interventions and supervisory relevance to be measured directly. Across five years of weekly DeFi exposure graphs and eight reference implementations, routing the LLM through forecasted evidence raises ticket F1 from 0.0076 for a conservative persistence-rules baseline to 0.0288 with Claude~Sonnet~4.6, the decision model we recommend (better F1 and explanations at $\sim$5$\times$ lower cost than Opus~4.7), but this buys coverage and auditability, not safety. The LLM over-reads the forecaster, misfiring on roughly $37\%$ of its interventions, and a stronger Opus~4.7 is no better ($44\%$, false-intervention rate $0.437$ even with the safety gate); over-intervention thus persists regardless of model, so safe high-severity action comes from the data-health and confidence gates and human review, not from the decision model. \sysname{} is thus an auditable recall-and-explanation option for human-in-the-loop DeFi supervision, not a replacement for conservative rule-based systems.

\section{Related Work}
\label{sec:related}

\paragraph{Bench positioning and ground-truth definition.}
LLM-agent benchmarks (HELM~\citep{liang2022helm},
SWE-bench~\citep{jimenez2024swebench},
AgentBench~\citep{liu2024agentbench}) score open-ended reasoning,
software repair, and generic agent behaviour; temporal-graph
benchmarks (TGB~\citep{huang2023tgb}, OGB~\citep{hu2020ogb}) score
structural prediction quality. Neither scores whether an LLM agent's
supervisory decisions match what a regulator would actually
prioritise. The closest prior systemic-risk evaluations
\citep{BERTOMEU2024105321,gonon2025computingsystemicriskmeasures,
li2025hgmae} rank protocols by \emph{fractional} weight change, which
disproportionately surfaces tiny protocols of low systemic relevance.
\benchname{} replaces this with an \emph{absolute-loss} ground
truth (\S\ref{sec:preliminaries}).

\paragraph{LLM agents in finance and DeFi.}
General agent patterns such as ReAct~\citep{YaoEtAl2023ReAct}
combine reasoning with tool use, and FinGPT~\citep{YangEtAl2023FinGPT}
adapts language models to financial data. DeFi-specific agents inherit
this template but, whatever their target (transaction
auditing~\citep{yao2025arbiter}, intent
mining~\citep{mao2025knowyourintent}, smart-contract
verification~\citep{hu2026heimdallr,kong2026knowdit},
price-manipulation detection~\citep{liu2025pmdetector}, anomaly
explanation~\citep{watson2025explainfirst}, asset-preference
auditing~\citep{Wu2026AuditingAssetPreferences}, or portfolio construction
from graph-plus-LLM encoders~\citep{luo2025llmcrypto,
jeon2026blindtrade}), all reason over \emph{raw} transactions or token
text and are judged on detection accuracy. None feeds an LLM
\emph{structured forecasted evidence}, and none reports a
false-intervention rate against a regulator-aligned ground truth, the
gap our characterisation (\S\ref{sec:results}) closes.

\paragraph{Forecast-grounded LLM agents in other domains.}
Pairing a domain forecaster with an LLM decision layer is an emerging
deployment pattern. In macroeconomics, ChatGPT-augmented PMI
nowcasting~\citep{deBondtSun2025ChatGPTNowcast} and LLM-driven
macroeconomic forecasting~\citep{CarrieroEtAl2024MacroLLM} both feed
structured numeric inputs into an LLM that produces interpretable
narratives; a BIS primer surveys the wider
pattern~\citep{KwonEtAl2024BISLLMprimer}. Time-series foundation
models such as Chronos~\citep{AnsariEtAl2024Chronos},
Lag-Llama~\citep{RasulEtAl2023LagLlama}, and
TimesFM~\citep{DasEtAl2024TimesFM} make the forecaster reusable
across tasks; tabular foundation models extend the same idea to
heterogeneous structured data~\citep{HollmannEtAl2025TabPFN,
eremeev2025turningtabularfoundationmodels}. \fmname{} is the closest
upstream forecaster for our setting: it adapts a graph-tabular foundation
model to dynamic DeFi credit-exposure graphs and exposes multi-horizon
forecasted networks for downstream measurement~\citep{ShuEtAl2026DeXposureFM}.
Our contribution is not the forecaster itself, but the agentic supervision
layer and evaluation harness that test whether those forecasts can ground
LLM-written supervisory tickets without hiding false interventions. None of
these forecast$\to$LLM pipelines, to our knowledge, has been scored against a
regulator-aligned ground truth for high-stakes financial-network supervision.

\section{Preliminaries}
\label{sec:preliminaries}

We model decentralized-finance credit exposure as a temporal weighted directed graph. Let $\mathcal{P}$ be the universe of protocols and $\mathcal{X}$ the universe of tokens. At time $t$, protocol $p$ holds tokens $X_p(t)\subseteq\mathcal{X}$, while protocol $q$ issues or is economically linked to tokens $X_G(q)\subseteq\mathcal{X}$. Protocol $p$ is exposed to $q$ if
$X_p(t)\cap X_G(q)\neq\emptyset$.
The exposure network is
$G_t=(V_t,E_t,W_t)$,
where $V_t\subseteq\mathcal{P}$ is the active protocol set, $E_t\subseteq V_t\times V_t$ is the directed exposure set, and $W_t[p,q]=w_{pq,t}\geq 0$ is the USD exposure from $p$ to $q$. For protocol $v$, define its incident exposure mass as
\begin{equation*}
w_t(v)=\sum_{u\in V_t}w_{uv,t}+\sum_{u\in V_t}w_{vu,t}.
\end{equation*}
The observed history is a sequence of snapshots $\mathcal{G}_{1:T}=\{G_1,\ldots,G_T\}$, optionally with covariates $X_t$.

At decision epoch $t$, the forecasting problem is to estimate $G_{t+h}$ for horizons $h\in\mathcal{H}$. A probabilistic temporal-graph forecaster outputs
\begin{equation*}
P_{t,h}=P(G_{t+h}\mid G_{\leq t},X_{\leq t}).
\end{equation*}
For each ordered pair $(p,q)$, let
\begin{gather*}
\pi_{pq,t+h}=\Pr(e_{pq,t+h}=1\mid G_{\leq t},X_{\leq t}),\\
\mu_{pq,t+h}
=
\mathbb{E}\!\left[
w_{pq,t+h}
\mid
e_{pq,t+h}=1,G_{\leq t},X_{\leq t}
\right].
\end{gather*}
A representative predicted graph $\widehat{G}_{t+h}$ is constructed by
\begin{gather*}
\widehat{E}_{t+h}
=
\{(p,q):\pi_{pq,t+h}>\pi_{\min}\}, \\
\widehat{w}_{pq,t+h}
=
\pi_{pq,t+h}\mu_{pq,t+h}.
\end{gather*}
Uncertainty is represented by Monte Carlo samples
\begin{equation*}
\widetilde{G}_{t+h}^{(1)},\ldots,\widetilde{G}_{t+h}^{(M)}
\sim P_{t,h}.
\end{equation*}

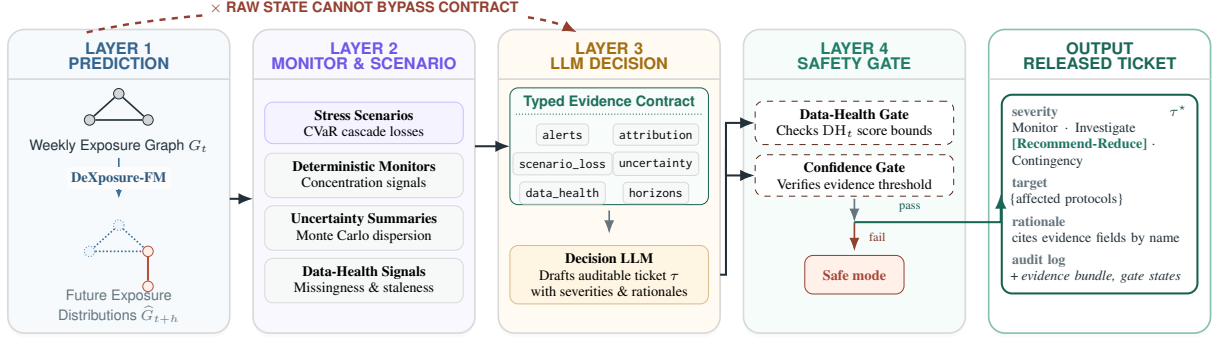
\begin{figure*}[t!]
\centering
\resizebox{\textwidth}{!}{

\definecolor{dxBInk}{HTML}{223139}%
\definecolor{dxBMuted}{HTML}{64757D}%
\definecolor{dxBRule}{HTML}{D6DEE2}%
\definecolor{dxBBlue}{HTML}{3B6F9E}%
\definecolor{dxBBlueBg}{HTML}{EFF6FB}%
\definecolor{dxBGreen}{HTML}{248A72}%
\definecolor{dxBGreenBg}{HTML}{EDF8F4}%
\definecolor{dxBGold}{HTML}{B47A16}%
\definecolor{dxBGoldBg}{HTML}{FFF5E3}%
\definecolor{dxBRed}{HTML}{B8503C}%
\definecolor{dxBRedBg}{HTML}{FFF0EA}%
\definecolor{dxBGrayBg}{HTML}{F7F8F8}%
\definecolor{dxBPurple}{HTML}{7B61FF}%
\definecolor{dxBPurpleBg}{HTML}{F5F3FF}%

\begin{tikzpicture}[
  font=\sffamily\small,
  bento/.style={
    draw=gray!30, fill=gray!2, rounded corners=8pt, line width=0.8pt, 
    drop shadow={opacity=0.04, shadow xshift=0pt, shadow yshift=-2pt}
  },
  bentotitle/.style={
    font=\sffamily\bfseries\fontsize{7.8pt}{8.4pt}\selectfont,
    align=center, text width=3.25cm
  },
  flow/.style={-{Latex[length=2.5mm,width=2.0mm]}, draw=dxBInk, line width=1.0pt},
  softflow/.style={-{Latex[length=2.0mm,width=1.5mm]}, draw=dxBMuted, line width=0.8pt},
  blocked/.style={-{Latex[length=2.5mm,width=2.0mm]}, draw=dxBRed!80!black, dashed, line width=1.2pt},
  itembox/.style={
    draw=gray!30, fill=white, rounded corners=4pt, 
    minimum width=3.4cm, minimum height=0.65cm, font=\scriptsize, align=center
  },
  contractbox/.style={
    draw=dxBGreen!70!black, fill=white, rounded corners=4pt, 
    minimum width=3.4cm, minimum height=2.0cm, font=\scriptsize, align=center, text=dxBInk
  },
  fieldpill/.style={
    fill=dxBGrayBg, draw=gray!30, rounded corners=2pt, font=\ttfamily\fontsize{6.5pt}{7pt}\selectfont, inner sep=3pt
  },
  node_curr/.style={circle, draw=dxBInk, fill=dxBInk!20, inner sep=1.8pt},
  node_futu/.style={circle, draw=dxBBlue, fill=dxBBlueBg, inner sep=1.8pt, densely dotted},
  node_futu_alert/.style={circle, draw=dxBRed, fill=dxBRedBg, inner sep=1.8pt},
  edge_curr/.style={draw=dxBInk, line width=0.8pt},
  edge_futu/.style={draw=dxBBlue, line width=0.8pt, densely dotted},
  edge_futu_alert/.style={draw=dxBRed, line width=1.0pt}
]

\def\colA{0}
\def\colB{4.2}
\def\colC{8.4}
\def\colD{12.6}
\def\colE{16.8}

\node[bento, fill=dxBBlueBg!55,   minimum width=3.8cm, minimum height=5.2cm] (P1) at (\colA,0) {};
\node[bento, fill=dxBPurpleBg!55, minimum width=3.8cm, minimum height=5.2cm] (P2) at (\colB,0) {};
\node[bento, fill=dxBGoldBg!45,   minimum width=3.8cm, minimum height=5.2cm] (P3) at (\colC,0) {};
\node[bento, fill=dxBGreenBg!50,  minimum width=3.8cm, minimum height=5.2cm] (P4) at (\colD,0) {};
\node[bento, fill=white, draw=dxBGreen!55, minimum width=3.8cm, minimum height=5.2cm] (P5) at (\colE,0) {};

\def\titleY{2.15}
\def\lineY{1.75}
\node[bentotitle, text=dxBBlue!90!black]   at (\colA, \titleY) {LAYER 1\\[-1pt]PREDICTION};
\node[bentotitle, text=dxBPurple!90!black] at (\colB, \titleY) {LAYER 2\\[-1pt]MONITOR \& SCENARIO};
\node[bentotitle, text=dxBGold!90!black]   at (\colC, \titleY) {LAYER 3\\[-1pt]LLM DECISION};
\node[bentotitle, text=dxBGreen!90!black]  at (\colD, \titleY) {LAYER 4\\[-1pt]SAFETY GATE};
\node[bentotitle, text=dxBGreen!75!black]  at (\colE, \titleY) {OUTPUT\\[-1pt]RELEASED TICKET};

\foreach \x in {\colA, \colB, \colC, \colD, \colE} {
    \draw[gray!30, line width=0.8pt] (\x-1.9, \lineY) -- (\x+1.9, \lineY);
}

\begin{scope}[shift={(\colA, 1.25)}]
    \node[node_curr] (n1) at (0, 0.25) {};
    \node[node_curr] (n2) at (-0.5, -0.2) {};
    \node[node_curr] (n3) at (0.5, -0.2) {};
    \draw[edge_curr] (n1) -- (n2); \draw[edge_curr] (n1) -- (n3); \draw[edge_curr] (n2) -- (n3);
\end{scope}
\node[font=\scriptsize\sffamily, text=dxBInk] at (\colA, 0.6) {Weekly Exposure Graph $G_t$};

\draw[-{Latex[length=1.5mm]}, dxBBlue, thick] (\colA, 0.4) -- (\colA, -0.3);
\node[font=\scriptsize\bfseries, fill=gray!2, inner sep=2pt, text=dxBBlue!80!black] at (\colA, 0.05) {\fmname};

\begin{scope}[shift={(\colA, -1.0)}]
    \node[node_futu] (f1) at (0, 0.25) {};
    \node[node_futu] (f2) at (-0.5, -0.2) {};
    \node[node_futu_alert] (f3) at (0.5, -0.2) {};
    \node[node_futu_alert] (f4) at (0.5, -0.8) {};
    \draw[edge_futu] (f1) -- (f2); \draw[edge_futu] (f1) -- (f3); \draw[edge_futu] (f2) -- (f3);
    \draw[edge_futu_alert] (f3) -- (f4);
\end{scope}
\node[font=\scriptsize\sffamily, text=dxBMuted, align=center] at (\colA, -2.15) {Future Exposure\\Distributions $\widehat{G}_{t+h}$};

\node[itembox, fill=dxBPurpleBg, draw=dxBPurple!40] at (\colB, 1.0) {\textbf{Stress Scenarios}\\CVaR cascade losses};
\node[itembox, fill=dxBGrayBg, draw=gray!30]        at (\colB, 0.1) {\textbf{Deterministic Monitors}\\Concentration signals};
\node[itembox, fill=dxBGrayBg, draw=gray!30]        at (\colB, -0.8) {\textbf{Uncertainty Summaries}\\Monte Carlo dispersion};
\node[itembox, fill=dxBGrayBg, draw=gray!30]        at (\colB, -1.7) {\textbf{Data-Health Signals}\\Missingness \& staleness};

\node[contractbox, fill=dxBGreenBg!40] (contract) at (\colC, 0.6) {};
\node[font=\sffamily\bfseries\scriptsize, text=dxBGreen!80!black] at (\colC, 1.35) {Typed Evidence Contract};
\draw[dxBGreen!70!black, line width=0.5pt, densely dotted] (\colC-1.6, 1.15) -- (\colC+1.6, 1.15);

\node[fieldpill] at (\colC-0.8, 0.8) {alerts};
\node[fieldpill] at (\colC+0.8, 0.8) {attribution};
\node[fieldpill] at (\colC-0.8, 0.3) {scenario\_loss};
\node[fieldpill] at (\colC+0.8, 0.3) {uncertainty};
\node[fieldpill] at (\colC-0.8, -0.2) {data\_health};
\node[fieldpill] at (\colC+0.8, -0.2) {horizons};

\draw[-Latex, thick, dxBMuted] (\colC, -0.5) -- (\colC, -0.9);

\node[itembox, fill=dxBGoldBg, draw=dxBGold!50, minimum height=1.0cm] (llm) at (\colC, -1.6) {\textbf{Decision LLM}\\Drafts auditable ticket $\tau$\\with severities \& rationales};

\node[itembox, fill=white, draw=dxBRed!50!black, dashed] (gate1) at (\colD, 1.0) {\textbf{Data-Health Gate}\\Checks $\operatorname{DH}_t$ score bounds};
\node[itembox, fill=white, draw=dxBRed!50!black, dashed] (gate2) at (\colD, 0.1) {\textbf{Confidence Gate}\\Verifies evidence threshold};

\draw[-Latex, thick, dxBMuted] (\colD, -0.3) -- (\colD, -0.7);
\draw[-Latex, thick, dxBRed!80!black] (\colD, -0.7) -- (\colD, -1.2);
\node[itembox, fill=dxBRedBg, draw=dxBRed!80!black, minimum width=1.7cm, minimum height=0.66cm, text=dxBRed!90!black, font=\scriptsize] (safe) at (\colD, -1.6) {\textbf{Safe mode}};
\node[font=\fontsize{6pt}{7pt}\selectfont, text=dxBRed!70!black, anchor=west] at (\colD+0.12,-0.95) {fail};

\node[draw=dxBGreen!70!black, fill=white, rounded corners=5pt, line width=1.0pt,
      anchor=north, text width=3.0cm, align=left, inner sep=5pt,
      font=\fontsize{6.8pt}{7.9pt}\selectfont, text=dxBInk] (tcard) at (\colE, 1.5) {%
  \hyphenpenalty=10000\exhyphenpenalty=10000\relax
  {\color{dxBMuted}\textbf{severity}} \hfill {\color{dxBGreen!70!black}$\tau^\star$}\\
  Monitor\, \textperiodcentered\, Investigate\\
  {\bfseries\color{dxBGreen!85!black}[Recommend-Reduce]} \textperiodcentered\, Contingency\\[3pt]
  {\color{dxBMuted}\textbf{target}}\\
  \{affected protocols\}\\[3pt]
  {\color{dxBMuted}\textbf{rationale}}\\
  cites evidence fields by name\\[3pt]
  {\color{dxBMuted}\textbf{audit log}}\\
  {\itshape +\,evidence bundle, gate states}
};

\draw[flow] (\colA+1.9, -0.3) -- (\colB-1.9, -0.3);
\draw[flow] (\colB+1.9, 0.6) -- (\colC-1.7, 0.6);

\draw[flow] (\colC+1.9, -1.6) -| (\colD-2.2, 0.55) |- (gate2.west);
\draw[flow] (\colD-2.2, 0.55) |- (gate1.west);

\draw[-{Latex[length=2.5mm,width=2.0mm]}, draw=dxBGreen!75!black, line width=1.0pt]
      (\colD, -0.7) -| (tcard.west);
\node[font=\fontsize{6pt}{7pt}\selectfont, text=dxBGreen!70!black, anchor=south] at (\colD+0.95,-0.66) {pass};

\draw[blocked] (\colA-0.55, 2.55) .. controls (\colA+1.65, 3.08) and (\colC-1.65, 3.08) .. (\colC-0.55, 2.55);
\node[fill=white, text=dxBRed!80!black, font=\bfseries\sffamily\scriptsize, inner sep=3pt] at (4.2, 2.98) {$\times$ RAW STATE CANNOT BYPASS CONTRACT};

\path[use as bounding box] (\colA-1.9,-2.62) rectangle (\colE+1.9,3.18);

\end{tikzpicture}}
\caption{\sysname{} system overview. A weekly exposure graph $G_t$ is
forecast into future exposure distributions and deterministic risk signals,
then converted into a typed evidence bundle containing monitor alerts,
attribution, stress-scenario losses, uncertainty summaries, data-health
signals, and forecast horizons. The LLM drafts supervisory tickets only from
this structured evidence, after which data-health and confidence gates either
release an auditable ticket $\tau^{\star}$ or enter safe mode.}
\label{fig:system}
\end{figure*}

A monitor maps graphs to scalar risk statistics $\Phi_j:\mathcal{G}\to\mathbb{R}$, such as maximum PageRank, HHI concentration, density, PageRank Gini, or degree Gini. An alert fires when the predicted-graph statistic $\Phi_j(\widehat{G}_{t+h})$ deviates from its length-$L$ rolling baseline by more than $z$ standard deviations; the rolling mean/standard deviation and the $z$-score alert rule are defined in Appendix~\ref{app:pipeline-math}.

A stress scenario is a graph perturbation $s:\mathcal{G}\to\mathcal{G}$ that removes exposure mass. We summarize its effect on the Monte Carlo samples by the conditional value-at-risk $\operatorname{CVaR}_{\lambda}(s,h)$, the mean of the worst $\lceil\lambda M\rceil$ sampled loss fractions; the per-scenario loss fraction $L(s;G)$ and $\operatorname{CVaR}$ are defined in Appendix~\ref{app:pipeline-math}.
Evaluation metrics are given in Appendix \ref{app:evaluation}.
\section{\sysname{} Pipeline}
\label{sec:pipeline}

This section introduces the \sysname{} Pipeline.

\subsection{Overview}

LLM-mediated supervision risks turning weak, stale, or uncertain evidence
into confident recommendations. \sysname{} therefore separates forecasting,
evidence building, ticket drafting, and ticket release
(Figure~\ref{fig:system}). Layer~1 forecasts future credit-exposure graphs
from the current weekly snapshot. Layer~2 builds a typed evidence bundle from
those forecasts: alerts, attribution, uncertainty estimates, and
stress-scenario losses. Layer~3 drafts ranked supervisory tickets from this
evidence alone, the only stage that calls the LLM. Layer~4 gates
intervention-level tickets before release.
This decomposition keeps the LLM as a constrained drafting component rather
than the release authority. Despite this four-stage decomposition, the
pipeline still runs at weekly cadence on a single
RTX~4090, with well under a dollar of LLM API spend per weekly decision (a
full \benchname{} leaderboard run is approximately \$10).

\subsection{Layer 1: Forecasting}
\label{ssec:forecaster}

At decision epoch $t$, \sysname{} observes the exposure graph $G_t$ defined in
Section~\ref{sec:preliminaries}. For each forecast horizon
$h\in\{1,4,8,12\}$ weeks, it queries \fmname{} for a predictive distribution
over future exposure and constructs an expected-weight predicted
graph $\widehat{G}_{t+h}$. It also draws Monte Carlo graph samples that carry
predictive uncertainty into Layer~2. The predicted-graph construction and all
forecast hyperparameters ($\pi_{\min}$, sample count, horizons) are given in
Appendix~\ref{app:fm-notation}.

\subsection{Layer 2: Monitoring}
\label{ssec:evidence}

Layer~2 turns the forecast into a typed evidence bundle along two parallel
tracks, monitors and stress scenarios. It also scores the snapshot's
\emph{data health} $\operatorname{DH}_t\in[0,1]$, the mean of freshness,
missingness, topology, and discrepancy checks
(Section~\ref{sec:preliminaries}), which later modulates both alert
confidence and the Layer~4 gates. Given $\widehat{G}_{t+h}$ and its
Monte Carlo samples, five
\emph{monitor functionals} summarize concentration, fragility, and
instability in the predicted graph: top-1 PageRank
\citep{BrinPage1998PageRank}, HHI concentration \citep{Rhoades1993HHI},
network density, PageRank Gini, and degree Gini. An alert fires when a
monitor statistic deviates from its rolling baseline by more than two
standard deviations. Each alert carries a marginal-contribution top-$K$
edge attribution and an uncertainty-aware confidence score
$C_t(a)\in[0,1]$, which decreases with lower data health, wider Monte Carlo
dispersion, and longer forecast horizons.

In parallel, a \emph{scenario engine} applies five standardized stress
shocks to both $\widehat{G}_{t+h}$ and each Monte Carlo sample: single
protocol failure, bridge cluster failure, stablecoin de-peg, sector lending
shock, and correlated top-$10$ stress. For each scenario $s$ and horizon
$h$, the engine aggregates the worst-half sampled losses as
$\operatorname{CVaR}_{0.5}(s,h)$. These alerts, attribution fields,
confidence scores, and scenario-CVaR table form the evidence consumed by the
LLM. Full monitor and scenario definitions are provided in
Appendix~\ref{app:pipeline-math}.

\subsection{Layer 3: Ticket Drafting}
\label{ssec:llm-decision}

Layer~3 is the only stage that invokes the LLM, which receives the evidence
bundle as a typed JSON payload containing
alerts with attribution top-$K$, a scenario-CVaR table, Monte Carlo
dispersion summaries, and the scalar data-health score
$\operatorname{DH}_t$. It drafts, but does not release, a ranked list of
supervisory tickets. Each ticket contains (a) a severity level from a fixed
four-action playbook (\texttt{Monitor}, \texttt{Investigate},
\texttt{Recommend-Reduce}, and \texttt{Contingency}), (b) a target set of
protocols, and (c) a rationale that cites specific evidence fields by name.
Each decision is drafted three times at temperature~$0$; the Jaccard overlap
of ticket targets and severities across the three drafts is the
\emph{target-stability} signal reported in our evaluation.

\subsection{Layer 4: Safety Gates}
\label{ssec:gates}

Candidate tickets are not released by default. Layer~4 applies data-health
and confidence gates to determine whether intervention-level recommendations
are feasible. The \emph{data-health gate} thresholds the Layer~2 score: if
$\operatorname{DH}_t<\tau_{\mathrm{data}}=0.7$, the system enters safe mode
and emits only \texttt{Monitor} or \texttt{Investigate} tickets. The
\emph{confidence gate} blocks intervention-level tickets whenever the mean
alert confidence satisfies $\bar{C}_t<\tau_{\mathrm{conf}}=0.6$, even if the
data-health gate passes. Every emitted ticket carries the full evidence
bundle, the LLM rationale, and the gate states that produced it. The
resulting audit log is the unit of release (Section~\ref{sec:lessons}).
The end-to-end procedure is given in Appendix~\ref{app:pipeline-math}
(Algorithm~\ref{alg:agent_loop}).

\section{Experiments}
\label{sec:evaluation}

This section describes experiments.

\subsection{Evaluation Questions}

We organise the evaluation around three questions. \textbf{RQ1}: does the
forecaster earn its place? \textbf{RQ2}: does forecast-grounding change the
agent? \textbf{RQ3}: is each component load-bearing? Answering all three means
scoring the forecast and the agent's interventions against a regulator-aligned
ground truth that standard leaderboards omit, so we build \benchname{}.

\subsection{\benchname{}}
\label{sec:bench}

We develop a \benchname{} harness that
scores the pipeline on six independent axes spanning
forecasting, warning, and decision quality (full schemas in
Table~\ref{tab:benchmark_schema}):
\begin{itemize}[nosep,leftmargin=1.3em,label=\textbullet]
\item \texttt{b1\_forecast} -- temporal graph forecast quality;
\item \texttt{b2\_warning} -- streaming early-warning lead time;
\item \texttt{b3\_calibration} -- predictive-uncertainty calibration;
\item \texttt{b4\_stress} -- what-if stress-scenario fidelity;
\item \texttt{b5\_decision} -- supervisory ticket quality;
\item \texttt{b6\_robustness} -- degraded-data robustness.
\end{itemize}

The decision and warning axes share the absolute-loss ground truth of
\S\ref{sec:preliminaries} (top-$\pi$ stressed set, $\pi\!=\!0.05$,
horizon $h\!=\!4$ weeks), avoiding fractional-change rankings that
over-weight tiny protocols; this yields
$|\mathcal{S}_t^h|\!\approx\!283$ protocols per week and $n\!=\!29$
scored test weeks, as the final four snapshots lack a 4-week-ahead
target.
The forecasting axes (\texttt{b1}, \texttt{b3}, \texttt{b6}) answer RQ1 by
scoring the forecast before any agent consumes it; the decision axis
(\texttt{b5}) answers RQ2 by holding the forecast fixed and varying only the
LLM's evidence, with the stress axis (\texttt{b4}) checking the upstream
scenario fidelity that decision grounding relies on; RQ3 ablates components on
\texttt{b5} under clean and degraded data.

\paragraph{Data, splits, and baselines.}
All experiments run on \corpusname{} weekly exposure graphs. The
corpus holds 4{,}300 protocols,
24{,}300 tokens, $43.7$M exposure entries, and $283$ snapshots from
2020-03 to 2025-08 \citep{wu2025dexposuredatasetbenchmarksinterprotocol}.
All model selection uses pre-2025 data, with 2025 held out as the
frozen test split (split dates in \S\ref{app:bench-details}).
The harness bundles eight reference methods (full specs in
Table~\ref{tab:reference_methods}):
\begin{itemize}[nosep,leftmargin=1.3em,label=\textbullet]
\item \texttt{h1} -- weighted-degree heuristic monitor;
\item \texttt{m1} -- persistence $+$ rules (decision baseline);
\item \texttt{m2} -- snapshot LLM, no forecast;
\item \texttt{m3} -- EvolveGCN~\citep{ParejaEtAl2020EvolveGCN}, a learned GNN;
\item \texttt{m4} -- \fmname{} only;
\item \texttt{m5} -- \fmname{} $+$ rules;
\item \texttt{m6} -- \fmname{} $+$ LLM (full stack);
\item \texttt{m7} -- \fmname{} $+$ LLM $+$ safety gate.
\end{itemize}

The predictor-only methods (\texttt{m3}, \texttt{m4}) emit no supervisory
tickets and are scored on the forecast axes only.
Full schemas and method details are in
\S\ref{app:bench-details}--\S\ref{app:reference-methods}; the
historical Terra/Luna, FTX, and SVB/USDC warning study is reported
separately from the 2025 leaderboard.


\subsection{Empirical Results}
\label{sec:results}

We answer each question as a verdict (headline numbers in
Tables~\ref{tab:rq1}--\ref{tab:rq3}). \textbf{RQ1}: persistence
predicts point values more accurately, but only the forecaster emits the trend
and calibrated uncertainty the agent consumes (\texttt{b1}, \texttt{b3},
\texttt{b6}).
\textbf{RQ2}: routing the LLM through forecast evidence rather than raw
snapshots buys coverage and auditability ($+31\%$ ticket F1 via a larger target set) but \emph{not}
per-target accuracy or safety; the decision layer misfires on
${\sim}37$--$44\%$ of its interventions across models. \textbf{RQ3}: every
gate and scenario component is load-bearing under clean or degraded data,
while the decision model is an efficiency, not a safety, choice.

\begin{table}[tb]
\centering\scriptsize
\setlength{\tabcolsep}{2.4pt}
\caption{\textbf{Forecast quality across predictors} (RQ1). Point
estimates ($n{=}29$, $h{=}4$; PR-MAE in $10^{-5}$), no week-level CIs.
Bold marks the per-column best among accurate predictors. $\S$~persistence
trend is $0$ by construction; $\dagger$~EvolveGCN's lower
$\Delta_{\mathrm{rel}}$ is weak-but-stable (not robustness), so unbolded;
PI-cov/ECE is delivered only by \fmname{}. Full per-horizon/-regime audits
in \S\ref{app:full-tables}.}
\label{tab:rq1}
\begin{tabular}{@{}lccccc@{}}
\toprule
Predictor ($h{=}4$) & PR-MAE$\downarrow$ & RankC.$\uparrow$ & Trend$\uparrow$ & $\Delta_{\mathrm{rel}}\downarrow$ & PI-cov/ECE \\
\midrule
persistence (\texttt{m1}) & \textbf{3.4} & \textbf{.570} & .000\textsuperscript{\S} & .148 & --- \\
EvolveGCN (\texttt{m3})   & 3.5 & .551 & .324 & .040\textsuperscript{\dag} & --- \\
\fmname{} (\texttt{m5})   & 4.5 & .558 & \textbf{.628} & \textbf{.113} & \textbf{.913/.013} \\
\bottomrule
\end{tabular}
\end{table}

\begin{table*}[!tb]
\centering\footnotesize
\setlength{\tabcolsep}{4pt}
\caption{\textbf{Ticket quality and matched-budget coverage} (RQ2). Frozen
2025 split ($n{=}29$); decisions Claude~Opus~4.7, judge Claude~Opus~4.8.
\emph{Interv.(wk)} = weeks with $\ge$1 intervention ticket;
\texttt{m1}/\texttt{m2}/\texttt{m5} never escalate, so their FIR is undefined
($0/0$, n/a), \emph{not} a safety guarantee. Brackets are $95\%$ bootstrap
CIs; $\dagger$\texttt{m1} from a local re-run; $\ast$\,saturated (no week
reaches depth $k$, so the cell repeats the full set). Panel (B) truncates each
method to its top-$k$ targets/week; significance tests and full panels in
\S\ref{app:stats},\,\S\ref{app:judge-panel}.}
\label{tab:rq2}
\begin{tabular*}{\textwidth}{@{\extracolsep{\fill}}llccccccc@{}}
\multicolumn{9}{@{}l}{\textbf{(A) End-to-end ticket quality}}\\
\toprule
ID & Stack & Interv.(wk) & Prec. & Rec. & F1$\uparrow$ [95\% CI] & FIR$\downarrow$ [95\% CI] & Stab. & Judge [CI] \\
\midrule
\texttt{m1} & persist.+rules & 0/29 & \textbf{.720} & .004 & .0076 [.004,.012]\textsuperscript{\dag} & n/a & .514 & --- \\
\texttt{m2} & snapshot LLM   & 0/29 & .575 & .009 & .0184 [.014,.022] & n/a & .532 & 2.24 [2.10,2.41] \\
\texttt{m5} & FM+rules       & 0/29 & .600 & .010 & .0190 & n/a & .257 & --- \\
\texttt{m6} & FM+LLM         & 17/29 & .570 & .012 & \textbf{.0241 [.020,.028]} & .448 [.293,.601] & .488 & 2.41 [2.21,2.62] \\
\texttt{m7} & FM+LLM+gate    & 16/29 & .580 & .012 & .0234 [.019,.027] & .437 [.282,.592] & .435 & \textbf{2.45 [2.24,2.66]} \\
\bottomrule
\end{tabular*}

\vspace{4pt}
\begin{tabular*}{\textwidth}{@{\extracolsep{\fill}}lcccccccc@{}}
\multicolumn{9}{@{}l}{\textbf{(B) Matched-budget} (R@$k\,{\times}10^{3}$ / P@$k$, top-$k$ targets/week)}\\
\toprule
Method & R@1 & P@1 & R@3 & P@3 & R@5 & P@5 & R@7 & P@7 \\
\midrule
\texttt{m1} persist.+rules & 0.71 & .24 & 2.45 & .26 & 4.05 & .26 & 4.05\textsuperscript{$\ast$} & .26 \\
\texttt{m2} snapshot LLM   & 1.53 & .52 & 5.70 & .59 & 9.08 & .58 & 9.35\textsuperscript{$\ast$} & .58 \\
\texttt{m6} FM+LLM         & 1.64 & .55 & 5.30 & .56 & 9.63 & .59 & \textbf{12.14} & .57 \\
\texttt{m7} FM+LLM+gate    & 1.64 & .55 & 5.30 & .56 & 9.13 & .57 & \textbf{11.95} & .58 \\
\bottomrule
\end{tabular*}
\end{table*}

\begin{table*}[!tb]
\centering\footnotesize
\setlength{\tabcolsep}{4pt}
\caption{\textbf{Component ablations and decision-model swap} (RQ3).
\emph{(A)} clean ablation vs \texttt{m5} (A1/A6 dormant on clean data).
\emph{(B)} data-health gate (A1) under degradation ($\tau_{\mathrm{conf}}{=}0$).
\emph{(C)} decision-model swap in \texttt{m7} ($95\%$ CIs). Full sweeps, A4/A5,
and crisis volumes ($\sim$4$\times$) in
\S\ref{app:a1-isolated},\,\S\ref{app:ablation-stress}.}
\label{tab:rq3}
\begin{tabular*}{\textwidth}{@{\extracolsep{\fill}}lcclccc@{}}
\toprule
\multicolumn{3}{@{}l}{\textbf{(A) Clean ablation} (vs \texttt{m5})} & \multicolumn{4}{l}{\textbf{(B) Degraded: data-health gate (A1)}}\\
\cmidrule(r){1-3}\cmidrule(l){4-7}
Config & Prec. & FIR & Regime$\,\cdot\,$gate & Prec. & FIR & \#int \\
\midrule
full \texttt{m5}        & .600 & n/a\textsuperscript{0/0} & extreme$\,\cdot\,$off & .490 & .510 & 57 \\
$-$ multi-horiz.\ (A6)  & .600 & $\to$App                  & extreme$\,\cdot\,$on  & .545 & \textbf{.000} & 0 \\
$-$ scenario (A3)       & \textbf{.000} & ---              & severe$\,\cdot\,$off  & .538 & .541 & 24 \\
$-$ data-health (A1)    & .600 & $\to$(B)                  & severe$\,\cdot\,$on   & .538 & .400 & 25 \\
$-$ conf.\ gate (A2)    & .600 & .429                      &                       &      &      &   \\
\bottomrule
\end{tabular*}

\vspace{4pt}
\begin{tabular*}{\textwidth}{@{\extracolsep{\fill}}lccc@{}}
\multicolumn{4}{@{}l}{\textbf{(C) Decision-model swap} (\texttt{m7}; efficiency, not safety)}\\
\toprule
Decision model & F1$\uparrow$ [95\% CI] & FIR$\downarrow$ [95\% CI] & rel.\ cost \\
\midrule
Claude Opus~4.7   & .0234 [.019,.027] & .437 [.282,.592] & 1$\times$ \\
Claude Sonnet~4.6 & \textbf{.0288 [.024,.034]} & .374 [.224,.529] & \textbf{$\sim$0.2$\times$} \\
Gemini~2.5~Pro    & .0139 [.011,.017] & \textbf{.190 [.069,.328]} & --- \\
\bottomrule
\end{tabular*}
\end{table*}

\paragraph{RQ1: forecaster vs.\ persistence.}
On static error, persistence is a deceptively strong baseline
(Table~\ref{tab:rq1}), leading on PageRank MAE and Spearman rank
correlation. \fmname{} instead contributes the signals persistence
structurally lacks, a non-trivial trend signal (persistence is $0$ by
construction), conformal calibration (PI coverage $.913$ at a $.90$ target,
ECE $.013$), and $24\%$ less degradation under data loss. EvolveGCN trails
persistence on accuracy, consistent with its $\sim$220 weekly-snapshot
training budget; full per-horizon and per-regime audits are in
\S\ref{app:full-tables}.

\paragraph{RQ2: grounding vs.\ raw snapshots.}
Routing the LLM through forecast evidence rather than raw snapshots lifts
ticket F1 by $+31\%$ (\texttt{m2}$\to$\texttt{m6}, $p\!<\!10^{-4}$;
Table~\ref{tab:rq2}A); this forecast-grounding gain and the LLM policy
alone (\texttt{m5}$\to$\texttt{m6}, $+27\%$) are additive and
non-substitutable, so the full agentic stack
(\texttt{m1}$\to$\texttt{m7}) reaches $+208\%$ over the persistence$+$rules
baseline. The baseline's higher single-axis precision (\texttt{m1}, $.720$)
is not a safety guarantee, since \texttt{m1}/\texttt{m2}/\texttt{m5} never
intervene at all. Adding the FM signal instead makes \texttt{m6} escalate,
and $44.8\%$ of those interventions misfire ($p\!<\!10^{-4}$); grounding is
$1.000$ for both variants, so the failure mode is not citation hallucination
but \emph{over-reading the forecaster's evidence as warrant for high-severity
action}. Two caveats temper the F1 story. \emph{(i)}~The explanation-quality
lift ($2.24\!\to\!2.45$ judge) is directional, not confirmed. It holds under a
GPT-5.5 judge but not the tier-above Claude~Opus~4.8
\citep{Zheng2023LLMJudge}, and cross-family judges agree only on ranking
\texttt{m7} top. \emph{(ii)}~At a matched budget the FM-fed and raw-snapshot
LLMs are indistinguishable per target up to the raw-snapshot model's ceiling
of about five targets per week (Table~\ref{tab:rq2}B, $k\!\le\!5$;
\S\ref{app:stats-budget}). The FM's contribution is not a higher per-target
hit rate but a \emph{larger} target set at undiminished precision. Past that
ceiling ($k\!=\!7$) it recovers significantly more stressed protocols at the
same precision ($p\!\le\!.0003$), and its tail targets hit at a rate
comparable to its head, so the extra targets are genuine detections rather
than padding. That expanded tail is also where over-intervention concentrates.

\paragraph{RQ3: component and model ablations.}
Read down the pipeline, each component owns one failure mode
(Table~\ref{tab:rq3}A). Multi-horizon forecasting (A6) and the scenario
engine (A3) supply the Layer~2 evidence; the data-health gate (A1) and
confidence gate (A2) guard the Layer~4 output. On clean data only two are
load-bearing. The scenario engine carries coverage, since skipping it
collapses ticket precision to $0$, and the confidence gate carries safety,
since removing it lifts FIR to $.429$ from the gated $0/0$. The other two
are dormant on clean data and act as reserves under stress, where the
data-health gate suppresses false alarms (FIR $[.27,.60]\!\to\![.00,.40]$,
and to $.000$ under the strict gate at $80$--$98\%$ feature/edge masking)
and multi-horizon monitoring raises crisis-window alert volume
$\sim$4$\times$ at unchanged precision (\S\ref{app:ablation-stress}).

Swapping the decision model trades cost for quality without touching safety
(Table~\ref{tab:rq3}C). A $\sim$5$\times$ cheaper Sonnet~4.6 raises
\texttt{m7} F1 over Opus~4.7 ($0.0234\!\to\!0.0288$; $p\!<\!0.001$) at
comparable intervention volume and the same over-intervention rate
(FIR $0.37$ vs $0.44$; n.s.), while Gemini~2.5~Pro lowers FIR only by
intervening far less at much lower F1 (\S\ref{app:stats}). Over-intervention
is thus bounded by the gates, not the model, so the deployment point is
\texttt{m7}$\times$Sonnet~4.6. The safety gate (\texttt{m6}$\to$\texttt{m7})
trades $3\%$ F1 within judge noise but adds a second auditable gate to every
intervention.

\section{Conclusion}
\label{sec:discussion-conclusion}
\label{sec:lessons}
\label{sec:conclusion}

We presented \sysname{}, a forecast-grounded agentic system for DeFi supervision. Its lesson is that LLM agents should not reason directly over raw on-chain data in high-stakes settings, since without structured evidence and escalation controls they over-read weak signals and over-intervene. \sysname{} instead routes decisions through \fmname{} forecasts, monitor and stress-scenario evidence, and data-health and confidence gates before emitting auditable tickets, while \benchname{}, our six-axis harness, scores them on ticket quality, regulator-aligned loss targeting, and false-intervention risk. Forecast-grounding improves coverage and auditability, but safe deployment hinges on false-intervention measurement and model calibration.

\section*{Limitations}
\label{sec:limitations}

\paragraph{Single domain.}
\benchname{} evaluates a single on-chain risk surface (DeFi
inter-protocol credit exposure). Generalisation to NFT lending,
perpetual derivatives, or cross-chain bridge networks is untested
and will require domain-specific re-calibration of the conformal
split and the stressed-set percentile $\pi$.

\paragraph{Weekly resolution.}
The pipeline operates at weekly granularity, but several DeFi shocks
unfold within hours; Terra/Luna erased \$40B in under 48~hours. A
sub-hourly production deployment would require denser snapshot
ingestion and re-evaluation of conformal calibration on the
resulting denser time series. The $4$--$5$ week lead times we
report should be read as ``how early the weekly monitor turns'',
not as end-to-end alert latency.

\paragraph{Pretraining exposure to named events.}
Well-known historical crises may appear in the decision and judge models'
pretraining data, so explanation-quality judgments on named events can be
inflated by recognition rather than evidence reading. We therefore report the
LLM-judge score only as a directional signal, and the headline capability
metrics do not depend on it. Ticket precision, F1, and false-intervention rate
are scored against the absolute-loss ground truth, citation grounding stays
traceable ($1.000$), and a fully anonymised crisis replay leaves only a small
target-precision change ($\Delta{=}.026$). The 2025 leaderboard split also
contains no comparably famous events. We defer the full masked-replay study to
the extended version.

\section*{Ethics Statement}
\label{sec:ethics}

\paragraph{Data provenance.}
\corpusname{} is built from public on-chain measurements
aggregated via DefiLlama~\citep{DefiLlama2026}. No private user data,
wallet-level personally-identifying information, or off-chain
identity data is used. Protocol-level aggregates are the unit of
analysis throughout.

\paragraph{Decision risk.}
\sysname{} emits supervisory tickets, not automated trades or
automated interventions. The high-severity action types
(\texttt{Recommend-Reduce}, \texttt{Contingency}) are gated on a
data-health score and a confidence score, and the pipeline emits a
safe-mode notice when either gate fails. We characterise the system
as decision support for a human supervisor, not as an autonomous
risk agent. The measured false-intervention rate
($\mathrm{FIR}\!\approx\!0.37$ for the deployed \fmname$+$Sonnet~4.6
variant, and not lowered by using a stronger model such as Opus~4.7)
is reported in the results tables so that downstream operators can
weigh the trade-off against their own risk tolerance before deployment.

\paragraph{LLM use disclosure.}
We used AI assistants to write code and to assist in writing this
paper.

\paragraph{Release.}
We will release the harness, the eight reference implementations,
and the ticket-level audit logs under a permissive open licence.
Model checkpoints will be released subject to dataset-licence
compatibility checks.

\bibliography{references}


\appendix
\makeatletter
\@addtoreset{algorithm}{section}
\@addtoreset{figure}{section}
\@addtoreset{table}{section}
\makeatother
\renewcommand{\thealgorithm}{\thesection.\arabic{algorithm}}
\renewcommand{\thefigure}{\thesection.\arabic{figure}}
\renewcommand{\thetable}{\thesection.\arabic{table}}
\section{Additional Algorithm Details}
\label{app:pipeline-math}

Section~\ref{sec:preliminaries} defines the exposure graph, forecast
distribution, predicted graph, monitor alert rule, scenario loss, ticket
metrics, and gate feasibility conditions used throughout the paper. This
appendix records only the concrete instantiation used by \sysname{}.

\subsection{Forecast instantiation}
\label{app:fm-notation}

At each decision epoch $t$, \sysname{} queries \fmname{} once for each
horizon $h\in\{1,4,8,12\}$ and constructs $\widehat{G}_{t+h}$ with the
expected-weight operator from Section~\ref{sec:preliminaries}. We set the
edge-existence threshold to $\pi_{\min}=0.2$, tuned on the validation split.
Uncertainty summaries use $M=50$ Monte Carlo graph samples. Each sample first
draws edge indicators from the predicted edge probabilities, then perturbs
positive edge weights with the forecaster's predictive regression spread. For
existing edges, residual predictions are folded into the conditional mean
$\mu_{pq,t+h}$ before applying the expected-weight construction; candidate
new edges enter only when their predicted existence probability exceeds
$\pi_{\min}$.

\subsection{Monitor instantiation}

\begin{table}[!htbp]
\centering\scriptsize
\caption{Monitor metrics. The five graph statistics tracked by the
structural monitor, each evaluated on the predicted graph
$\widehat{G}_{t+h}$.}
\label{tab:monitor-metrics}
\begin{tabular}{@{}lll@{}}
\toprule
ID & Metric                & Definition \\
\midrule
N1 & Systemic Importance  & $\max_{p\in V}\mathrm{PageRank}(\widehat{G}_{t+h})_p$ \\
N2 & HHI Concentration    & $\sum_p (d_p / \sum_q d_q)^2$, $d_p=\sum_q\widehat{w}_{pq,t+h}$ \\
N3 & Network Density      & $|\widehat{E}_{t+h}| / |V|(|V|-1)$ \\
N4 & PageRank Gini        & $\mathrm{Gini}(\{\mathrm{PageRank}_p\}_{p\in V})$ \\
N5 & Degree Gini          & $\mathrm{Gini}(\{d_p\}_{p\in V})$ \\
\bottomrule
\end{tabular}
\end{table}

Given a rolling window of length $L$, each monitor $\Phi_j$ maintains a
baseline mean and standard deviation
\begin{gather*}
\mu_{j,t}=\frac{1}{L}\sum_{\ell=1}^{L}\Phi_j(G_{t-\ell}),\\
\sigma_{j,t}=\sqrt{\frac{1}{L-1}\sum_{\ell=1}^{L}\left(\Phi_j(G_{t-\ell})-\mu_{j,t}\right)^2},
\end{gather*}
and an alert fires when the predicted-graph statistic leaves the band,
\begin{equation*}
A_{j,t,h}=\mathbf{1}\!\left\{\frac{\left|\Phi_j(\widehat{G}_{t+h})-\mu_{j,t}\right|}{\max(\sigma_{j,t},\epsilon)}>z\right\}.
\end{equation*}
We instantiate this rule with a $42$-week baseline, $z=2$, and
$\epsilon=10^{-8}$. PageRank uses damping $0.85$ and $10$ power iterations.
Gini coefficients are computed over active nodes in the predicted graph.

\subsection{Scenario instantiation}

\begin{table}[b]
\centering\small
\caption{Stress scenarios. The five what-if shocks applied to the
predicted graph, with the targeted node set and shock magnitude of each.}
\label{tab:stress-scenarios}
\begin{tabular}{@{}llll@{}}
\toprule
ID & Scenario               & Target                 & Shock \\
\midrule
S1 & Single protocol fail   & Highest-weight node    & 100\% \\
S2 & Bridge cluster fail    & All bridge nodes       & 100\% \\
S3 & Stablecoin de-peg      & All stablecoin nodes   & 50\%  \\
S4 & Sector lending shock   & All lending nodes      & 30\%  \\
S5 & Correlated stress      & Top-10 by weight       & 20\%  \\
\bottomrule
\end{tabular}
\end{table}

Each shock maps a graph to a perturbed graph $G^s$ with loss fraction
\begin{gather*}
L(s;G)=\frac{\operatorname{Mass}(G)-\operatorname{Mass}(G^s)}{\operatorname{Mass}(G)},\\
\operatorname{Mass}(G)=\sum_{e\in E(G)}w_e .
\end{gather*}
For the sampled graphs let $\ell_m(s,h)=L(s;\widetilde{G}_{t+h}^{(m)})$ with
sorted losses $\ell_{(1)}\geq\cdots\geq\ell_{(M)}$; the conditional
value-at-risk is
\begin{equation*}
\operatorname{CVaR}_{\lambda}(s,h)=\frac{1}{\lceil\lambda M\rceil}\sum_{m=1}^{\lceil\lambda M\rceil}\ell_{(m)}.
\end{equation*}
Scenario targets use protocol-sector metadata from \corpusname{}. For each
scenario and horizon, \sysname{} reports the deterministic loss on
$\widehat{G}_{t+h}$ and the sampled $\operatorname{CVaR}_{0.5}$.

\subsection{Gate instantiation and ticket scoring}

The data-health score of Section~\ref{sec:preliminaries} averages four
data-quality checks,
\begin{equation*}
\operatorname{DH}_t
=
\frac{1}{4}
\left(
\operatorname{fresh}_t+
\operatorname{miss}_t+
\operatorname{topo}_t+
\operatorname{disc}_t
\right),
\end{equation*}
and the data-health gate fires at $\tau_{\mathrm{data}}=0.7$. Safe mode still releases monitor alerts and
scenario summaries, but suppresses intervention-level tickets
(\textsc{RecommendReduce} and \textsc{Contingency}). The confidence gate uses
$\tau_{\mathrm{conf}}=0.6$ and clamps each Monte Carlo coefficient of
variation to $[0,1]$ before computing alert confidence
\begin{equation*}
C_t(a)=\operatorname{DH}_t\cdot\frac{1}{1+\operatorname{CV}_{t,h}(a)}\cdot\frac{1}{1+\log(1+h)},
\end{equation*}
where $\operatorname{CV}_{t,h}(a)$ is the Monte Carlo coefficient of variation
of alert $a$ at horizon $h$.

Feasible tickets are ranked by
\begin{equation}
\mathrm{Score}_t(u)
=
w_u\cdot \bar{C}_t\cdot \mathrm{Imp}_t,
\end{equation}
where $w_u\in\{0.25,0.50,0.75,1.00\}$ is the playbook severity weight for
the four ordered actions and $\mathrm{Imp}_t$ is the mean scenario CVaR
(default $1.0$ when no scenario summary is available).

\subsection{End-to-end procedure}
Algorithm~\ref{alg:agent_loop} ties the components above together into a
single decision epoch. Given the current graph snapshot and features, the
system first computes a data-health score and sets the safe-mode flag, then
for each horizon produces a forecast, builds the predicted graph, draws Monte
Carlo samples, derives alerts with confidences, and evaluates scenario losses.
The playbook stage finally maps the resulting alerts, scenario summary, and
data-health state into tickets.

\begin{algorithm}[t]
\caption{\sysname{}: one decision epoch.}
\label{alg:agent_loop}
\begin{algorithmic}[1]
\Require $(G_t,X_t)$; horizons $\mathcal{H}$; scenarios
$\mathcal{S}_0$; thresholds
$(\tau_{\mathrm{data}},\tau_{\mathrm{conf}},z,\pi_{\min})$; MC count $M$
\Ensure Alerts $\mathcal{A}_t$; scenario summary $\mathcal{R}_t$;
tickets $\mathcal{D}_t$
\State $\operatorname{DH}_t\gets\mathrm{DataHealth}(G_t,X_t)$
\State $\mathrm{SAFE\_MODE}_t\gets
       \mathbb{I}\{\operatorname{DH}_t<\tau_{\mathrm{data}}\}$
\ForAll{$h\in\mathcal{H}$}
  \State $P_{t,h}\gets\text{\fmname.Forecast}(G_t,X_t,h)$
  \State $\widehat{G}_{t+h}\gets
         \mathrm{BuildPredGraph}(P_{t,h};\pi_{\min})$
  \State $\{\widetilde{G}^{(m)}_{t+h}\}_{m=1}^M\gets
         \mathrm{MCSample}(P_{t,h})$
  \State Compute monitor statistics on $\widehat{G}_{t+h}$, attach rolling
         baselines, and derive alerts $\mathcal{A}_{t,h}$
  \State Compute alert confidences $C_t(\cdot)$ from sampled monitor
         dispersion
  \ForAll{$s\in\mathcal{S}_0$}
    \State Compute deterministic loss $L(s;\widehat{G}_{t+h})$ and sampled
           $\mathrm{CVaR}_{0.5}(s,h)$
  \EndFor
\EndFor
\State $\mathcal{D}_t\gets\mathrm{PlaybookDecision}($
\Statex \hspace{\algorithmicindent}
        $\mathcal{A}_t,\mathcal{R}_t,\operatorname{DH}_t,$
\Statex \hspace{\algorithmicindent}
        $\mathrm{SAFE\_MODE}_t,\tau_{\mathrm{conf}})$
\State \Return $(\mathcal{A}_t,\mathcal{R}_t,\mathcal{D}_t)$
\end{algorithmic}
\end{algorithm}

\section{Additional Implementation Details}
\label{app:implementation}

\subsection{Evaluation}
\label{app:evaluation}

For evaluation, we define regulator-aligned ground truth by absolute exposure loss. For protocol $v$ and horizon $h$,
\begin{equation}
\Delta_t^h(v)=w_t(v)-w_{t+h}(v).
\end{equation}
Given percentile $\pi\in(0,1)$, the stressed set is
\begin{equation}
S_t^h
=
\operatorname{top}_{\pi}
\left\{
v:
w_t(v)>0,\,
\Delta_t^h(v)>0
\right\},
\end{equation}
where protocols are ranked by $\Delta_t^h(v)$.

A supervisory system emits tickets
\begin{equation}
D_t=\{\tau_{t,1},\ldots,\tau_{t,K_t}\},
\qquad
\tau_{t,k}=(u_{t,k},Y_{t,k},r_{t,k}),
\end{equation}
where $u_{t,k}$ is an action, $Y_{t,k}\subseteq V_t$ is a target set, and $r_{t,k}$ is a rationale. The action space is
\begin{align*}
\mathcal{U}
=
\{&
\textsc{Monitor},
\textsc{Investigate},\\
&\textsc{RecommendReduce},
\textsc{Contingency}
\}.
\end{align*}
For predicted targets $\widehat{S}_t$, ticket quality is
\begin{gather*}
\operatorname{Precision}
=
\frac{|\widehat{S}_t\cap S_t^h|}{|\widehat{S}_t|},\\
\operatorname{Recall}
=
\frac{|\widehat{S}_t\cap S_t^h|}{|S_t^h|},\\
\operatorname{F1}
=
\frac{
2\,\operatorname{Precision}\,\operatorname{Recall}
}{
\operatorname{Precision}+\operatorname{Recall}
}.
\end{gather*}
Let $\widehat{S}^{\mathrm{int}}_t$ be targets of intervention-level actions in $\{\textsc{RecommendReduce},\textsc{Contingency}\}$. The false-intervention rate is
\begin{equation}
\operatorname{FIR}
=
\frac{
|\widehat{S}^{\mathrm{int}}_t\setminus S_t^h|
}{
|\widehat{S}^{\mathrm{int}}_t|
}.
\end{equation}

Finally, the data-health score $\operatorname{DH}_t\in[0,1]$ averages four
data-quality checks (freshness, missingness, topology, discontinuity); its
exact form is in Appendix~\ref{app:pipeline-math}. For alert $a$ at horizon $h$, the confidence $C_t(a)\in[0,1]$ combines data health, Monte Carlo dispersion, and forecast horizon, decreasing as data health falls or dispersion and horizon grow (exact form in Appendix~\ref{app:pipeline-math}). With $\bar{C}_t$ the mean active-alert confidence, intervention-level actions are feasible only when $\operatorname{DH}_t\geq\tau_{\mathrm{data}}$ and $\bar{C}_t\geq\tau_{\mathrm{conf}}$.

\subsection{Forecaster architecture and training}
\label{ssec:fm-spec}
\fmname{}~\citep{ShuEtAl2026DeXposureFM} is built on the GraphPFN graph-tabular foundation
backbone~\citep{HollmannEtAl2025TabPFN,
eremeev2025turningtabularfoundationmodels} with a LiMiX-16M tabular
transformer encoder ($12$ layers, $4$ attention heads, hidden dimension
$64$). The same backbone serves three heads (edge existence, binary;
edge weight, regression; and node-TVL change, regression), trained
jointly with a BCE$+$MSE composite loss. We fine-tune one checkpoint per
horizon $h\!\in\!\{1,4,8,12\}$ for $20$ epochs with AdamW (learning rate
$10^{-4}$ for heads, $10^{-5}$ for the backbone). Fine-tuning draws on
\corpusname{} snapshots from March~2020 to January~2025 ($104$ training
weeks, $12$ validation weeks for early stopping, $8$ held-out weeks for
internal evaluation); all data that influences weights or checkpoint
selection precedes the frozen 2025 leaderboard split
(\S\ref{app:bench-details}). Full-graph in-context inference fits a single
RTX~4090, and one weekly forecast across all four horizons completes in
well under a minute. These intrinsic forecaster numbers are kept out of
the main text because the paper's contribution is the operating point of
the full pipeline, not the FM's headline accuracy.

\subsection{Benchmark details}
\label{app:bench-details}

\paragraph{Six-axis schema.}
\begin{table*}[t]
\centering
\footnotesize
\setlength{\tabcolsep}{4pt}
\caption{Six-axis benchmark schema. Each axis is evaluated independently so
systems can be compared at the predictor, monitor, scenario, and decision
layers without re-running the full suite.}
\label{tab:benchmark_schema}
\begin{tabularx}{\textwidth}{@{}l l >{\raggedright\arraybackslash}X@{}}
\toprule
ID                       & Capability tested                & Primary metrics \\
\midrule
\texttt{b1\_forecast}    & Temporal graph prediction        & PageRank/HHI/Gini MAE, trend consistency, Spearman; per $h\!\in\!\{1,4,8,12\}$ \\
\texttt{b2\_warning}     & Streaming anomaly detection      & Precision, recall, F1, lead time, alert stability; per event $\times$ budget \\
\texttt{b3\_calibration} & Predictive uncertainty           & PI coverage at $0.90$ target, PI width, ECE, CRPS \\
\texttt{b4\_stress}      & What-if scenario fidelity        & Loss MAE, distressed-count MAE, propagation depth MAE, target overlap@$k$ \\
\texttt{b5\_decision}    & Supervisory ticket quality       & Precision, recall@$k$, F1, target stability, judge score, FIR \\
\texttt{b6\_robustness}  & Data quality sensitivity         & Per-benchmark metrics under 5 degradation regimes; relative degradation \\
\bottomrule
\end{tabularx}
\end{table*}
The six axes are non-overlapping (a system can excel at \texttt{b1} while
failing \texttt{b5}), decomposable along the four-layer framework so that
predictor (\texttt{b1}, \texttt{b3}, \texttt{b6}), monitor (\texttt{b2}),
scenario (\texttt{b4}), and decision (\texttt{b5}) ablations report
independently, and individually publishable so downstream work may extend a
single axis without re-running the suite.

\paragraph{Dataset split.}
\begin{table}[!htbp]
\centering\small
\caption{Dataset split. Temporal train/validation/test partition of the
weekly exposure graphs and the role of each split.}
\label{tab:dataset-split}
\begin{tabularx}{\linewidth}{@{}llrX@{}}
\toprule
Split      & Period             & Weeks       & Usage \\
\midrule
Train      & 2020-03--2024-06   & $\sim$222   & Forecaster training \\
Validation & 2024-07--2024-12   & $\sim$26    & Conformal calibration, threshold tuning \\
Test       & 2025-01--2025-08   & $\sim$33    & Frozen leaderboard \\
\bottomrule
\end{tabularx}
\end{table}

\paragraph{Historical warning windows.}
The \texttt{b2\_warning} event study is separate from the frozen 2025
leaderboard split. It evaluates the shared weighted-degree monitor around
Terra/Luna (2022-05-09), FTX (2022-11-07), and SVB/USDC (2023-03-10) with a
shorter $26$-week rolling baseline, chosen so each event window has enough
pre-event history. The 2025 leaderboard runs use the $42$-week monitor
baseline in Appendix~\ref{app:pipeline-math}.

\paragraph{Ground-truth definition.}
With $w_t(v)\!=\!\sum_{e\ni v}w_t(e)$ the total weight incident on $v$, the
regulator-aligned stressed set at horizon $h$ is
\begin{equation}
\begin{aligned}
\Delta_t^h(v) &= w_t(v)-w_{t+h}(v),\\
\mathcal{S}_t^h &=
\mathrm{top}_{\pi}\{v:\, w_t(v)>0,\ \Delta_t^h(v)>0\},
\end{aligned}
\label{eq:stressed-app}
\end{equation}
with $\pi\!=\!0.05$ and ground-truth horizon $h\!=\!4$~weeks. The final
four test weeks lack the $4$-week-ahead snapshot $w_{t+4}$ and are not
scored, leaving $n\!=\!29$ evaluated weeks. This is the single ground-truth
definition used by \texttt{b2}, \texttt{b5}, and the LLM-decision pipeline.
A submission produces one JSON file per \{benchmark,~method\} pair plus a
single \texttt{results.json}; the harness fixes random seeds, snapshots
library versions, and emits SHA-256 hashes of all checkpoint files.

\subsection{Reference methods}
\label{app:reference-methods}
\begin{table*}[t]
\centering
\footnotesize
\setlength{\tabcolsep}{4pt}
\caption{Reference methods used in the benchmark and ablations. The methods
cover predictor, monitor, decision-layer, and safety-gate variants.}
\label{tab:reference_methods}
\begin{tabularx}{\textwidth}{@{}l l l >{\raggedright\arraybackslash}X@{}}
\toprule
ID                              & Predictor      & Policy                & Role \\
\midrule
\texttt{h1\_weighted\_degree}   & ---            & heuristic alert       & Streaming anomaly baseline (\texttt{b2}) \\
\texttt{m1\_persistence\_rules} & $G_{t+h}\!=\!G_t$ & rule engine        & Decision baseline \\
\texttt{m2\_snapshot\_llm}      & current snap   & LLM                   & LLM-only ablation (no forecast) \\
\texttt{m3\_evolvegcn}          & EvolveGCN      & ---                   & Learned-GNN baseline \\
\texttt{m4\_fm\_only}           & \fmname{}      & ---                   & FM-only ablation \\
\texttt{m5\_fm\_rules}          & \fmname{}      & rule engine           & FM + rules \\
\texttt{m6\_fm\_llm}            & \fmname{}      & LLM                   & Full FM + LLM stack \\
\texttt{m7\_fm\_llm\_gated}     & \fmname{}      & LLM + safety gate     & Safety-gated FM + LLM \\
\bottomrule
\end{tabularx}
\end{table*}
These methods expose all four cuts of the four-layer ablation: adding the
predictor (\texttt{m1}$\!\to\!$\texttt{m5}), swapping the decision layer
(\texttt{m5}$\!\to\!$\texttt{m6}), adding the safety gate
(\texttt{m6}$\!\to\!$\texttt{m7}), and removing the predictor from the LLM
stack (\texttt{m2}$\!\leftrightarrow\!$\texttt{m6}). All conform to a
shared \texttt{predict\_graph}\,/\,\texttt{decide\_actions} interface.

\paragraph{LLM model panel and decoding.}
All LLM-bearing methods (\texttt{m2}, \texttt{m6}, \texttt{m7}) are
evaluated under three decision-layer configurations: Claude~Opus~4.7
(main), Claude~Sonnet~4.6 (within-family, tier-below ablation), and
Google~Gemini~2.5~Pro (cross-family ablation). Each runs at temperature $0$
and $\max\_tokens=4096$; self-consistency is probed with three repeated
calls per week and Jaccard overlap on emitted ticket targets and
severities. The LLM-as-judge for the explanation-quality axis is a
three-model panel chosen so the judge tier is at or above the decision
tier: Claude~Opus~4.8 (strictly tier-above Opus~4.7),
Google~Gemini~2.5~Pro, and OpenAI~GPT-5.5; the Gemini and GPT judges run
with OpenRouter's $\mathrm{reasoning}=\{\mathrm{effort:minimal}\}$ so the
judge thinking budget matches the non-reasoning Anthropic baseline.
Claude~Haiku~4.5 is reported as a weak-judge reference. All models are
served through OpenRouter; observed cost is on the order of a few US
dollars per full sweep of the 29-week test split.

\section{Additional Experimental Results}
\label{app:full-tables}

This appendix expands the headline numbers of \S\ref{sec:results} into the
per-slice audits underlying them, then reports week-level uncertainty for
the central comparisons.

\subsection{Per-slice result tables}

\paragraph{Full all-horizon \texttt{b1\_forecast} results.}
\label{app:b1-full}
\begin{table*}[!htbp]
\centering
\footnotesize
\setlength{\tabcolsep}{4pt}
\caption{\texttt{b1\_forecast} all-horizon forecasting on the 2025 test split.
PageRank MAE shown in $10^{-5}$ units. Best per horizon and metric in
\textbf{bold}; ties share the bold. Trend consistency for
\texttt{m1\_persistence} is structurally $0$ because
$\hat G_{t+h}=G_t$ produces no trend signal.}
\label{tab:b1_all_horizons}
\begin{tabular}{l l c c c c c}
\toprule
$h$ & Method & PR MAE ($\downarrow$) & HHI MAE ($\downarrow$) & Gini MAE ($\downarrow$) & Rank Corr ($\uparrow$) & Trend Cons ($\uparrow$) \\
\midrule
\multirow{4}{*}{1}
& \texttt{m1\_persistence} & \textbf{3.6} & \textbf{0.0376} & \textbf{0.0027} & \textbf{0.556} & 0.000 \\
& \texttt{m3\_evolvegcn}   & 3.7          & 0.0481 & 0.1149          & 0.535          & 0.444 \\
& \texttt{m4\_fm\_only}    & 4.7          & \textbf{0.0376} & 0.0029          & 0.549          & \textbf{0.531} \\
& \texttt{m5\_fm\_rules}   & 4.7          & \textbf{0.0376} & 0.0029          & 0.549          & 0.525 \\
\midrule
\multirow{4}{*}{4}
& \texttt{m1\_persistence} & \textbf{3.4} & 0.0322 & \textbf{0.0018} & \textbf{0.570} & 0.000 \\
& \texttt{m3\_evolvegcn}   & 3.5          & 0.0489 & 0.1326          & 0.551          & 0.324 \\
& \texttt{m4\_fm\_only}    & 4.5          & \textbf{0.0321} & 0.0022 & 0.559          & 0.621 \\
& \texttt{m5\_fm\_rules}   & 4.5          & \textbf{0.0321} & 0.0022 & 0.558          & \textbf{0.628} \\
\midrule
\multirow{4}{*}{8}
& \texttt{m1\_persistence} & \textbf{3.9} & 0.0399 & \textbf{0.0028} & \textbf{0.517} & 0.000 \\
& \texttt{m3\_evolvegcn}   & 4.0          & 0.0492 & 0.1140          & 0.499          & 0.296 \\
& \texttt{m4\_fm\_only}    & 4.5          & \textbf{0.0398} & 0.0030 & 0.508          & 0.624 \\
& \texttt{m5\_fm\_rules}   & 4.5          & \textbf{0.0398} & 0.0030 & 0.508          & \textbf{0.632} \\
\midrule
\multirow{4}{*}{12}
& \texttt{m1\_persistence} & \textbf{3.7} & 0.0393 & \textbf{0.0027} & \textbf{0.525} & 0.000 \\
& \texttt{m3\_evolvegcn}   & 3.8          & 0.0534 & 0.1245          & 0.509          & 0.314 \\
& \texttt{m4\_fm\_only}    & 4.1          & \textbf{0.0392} & 0.0029 & 0.516          & \textbf{0.600} \\
& \texttt{m5\_fm\_rules}   & 4.1          & \textbf{0.0392} & 0.0029 & 0.516          & \textbf{0.600} \\
\bottomrule
\end{tabular}
\end{table*}

Across all four horizons the persistence baseline (\texttt{m1}) stays best
on PageRank and Gini MAE and on rank correlation, while the FM-grounded
forecasters (\texttt{m4}/\texttt{m5}) lead on HHI MAE and trend consistency;
the gaps are stable as $h$ grows from $1$ to $12$.

\paragraph{Per-scenario \texttt{b4\_stress} results.}
\label{app:b4-detail}
\begin{table*}[!htbp]
\centering
\footnotesize
\setlength{\tabcolsep}{4pt}
\caption{\texttt{b4\_stress} per-scenario detail. Loss MAE in normalised
units; Distress count MAE in protocols; target overlap@10 measured against
the actual future graph. Best per scenario and metric in \textbf{bold};
ties share the bold, and columns where all four methods tie are left
unbolded. The main pattern is conservative: FM variants roughly match the
strong persistence baseline on loss and target-overlap metrics, while
EvolveGCN is generally weaker on downstream what-if fidelity.}
\label{tab:b4_detail}
\begin{tabular}{l l c c c c}
\toprule
Scenario & Method & Loss MAE ($\downarrow$) & Distress MAE ($\downarrow$) & Prop.~Depth MAE ($\downarrow$) & Overlap@10 ($\uparrow$) \\
\midrule
\multirow{4}{*}{S1 (single protocol)}
& \texttt{m1\_persistence} & 0.0668 & \textbf{200.3} & \textbf{0.897} & \textbf{0.314} \\
& \texttt{m3\_evolvegcn}   & 0.1072 & 218.1 & 0.966 & 0.034 \\
& \texttt{m4\_fm\_only}    & \textbf{0.0667} & 201.4 & \textbf{0.897} & \textbf{0.314} \\
& \texttt{m5\_fm\_rules}   & \textbf{0.0667} & 201.6 & \textbf{0.897} & \textbf{0.314} \\
\midrule
\multirow{4}{*}{S2 (bridge cluster)}
& \texttt{m1\_persistence} & 0.0150 & \textbf{27.3} & 0.000 & 0.521 \\
& \texttt{m3\_evolvegcn}   & \textbf{0.0116} & 196.4 & 0.000 & 0.189 \\
& \texttt{m4\_fm\_only}    & 0.0151 & 50.3 & 0.000 & \textbf{0.525} \\
& \texttt{m5\_fm\_rules}   & 0.0151 & 52.1 & 0.000 & \textbf{0.525} \\
\midrule
\multirow{4}{*}{S3 (stablecoin de-peg)}
& \texttt{m1\_persistence} & \textbf{0.0079} & 0.0 & 0.000 & \textbf{0.445} \\
& \texttt{m3\_evolvegcn}   & 0.0156 & 0.0 & 0.000 & 0.269 \\
& \texttt{m4\_fm\_only}    & 0.0080 & 0.0 & 0.000 & \textbf{0.445} \\
& \texttt{m5\_fm\_rules}   & 0.0080 & 0.0 & 0.000 & \textbf{0.445} \\
\midrule
\multirow{4}{*}{S4 (sector lending)}
& \texttt{m1\_persistence} & 0.0100 & 0.0 & 0.034 & \textbf{0.536} \\
& \texttt{m3\_evolvegcn}   & 0.0200 & 0.0 & 0.034 & 0.292 \\
& \texttt{m4\_fm\_only}    & \textbf{0.0099} & 0.0 & 0.034 & \textbf{0.536} \\
& \texttt{m5\_fm\_rules}   & \textbf{0.0099} & 0.0 & 0.034 & \textbf{0.536} \\
\midrule
\multirow{4}{*}{S5 (correlated top-10)}
& \texttt{m1\_persistence} & 0.0100 & 0.0 & \textbf{0.000} & \textbf{0.602} \\
& \texttt{m3\_evolvegcn}   & 0.0538 & 0.0 & 1.000 & 0.330 \\
& \texttt{m4\_fm\_only}    & \textbf{0.0099} & 0.0 & \textbf{0.000} & \textbf{0.602} \\
& \texttt{m5\_fm\_rules}   & \textbf{0.0099} & 0.0 & \textbf{0.000} & \textbf{0.602} \\
\bottomrule
\end{tabular}
\end{table*}

Across the five stress scenarios the FM variants (\texttt{m4}/\texttt{m5})
track the persistence baseline on loss MAE and target overlap@10, while
EvolveGCN (\texttt{m3}) is the weakest on downstream what-if fidelity.

\paragraph{Per-regime \texttt{b6\_robustness} results.}
\label{app:b6-detail}
\begin{table*}[!htbp]
\centering
\footnotesize
\setlength{\tabcolsep}{4pt}
\caption{\texttt{b6\_robustness} per-regime detail. All metrics at $h{=}4$
on the 2025 test split under five degradation regimes. PageRank, HHI, and
Gini MAE in $10^{-3}$ units. Relative degradation is signed against the
clean \texttt{b1\_forecast} baseline; negative values mean the predictor is
no worse than on clean data. Best per metric (lowest MAE, highest rank
correlation, smallest $|\Delta_{\mathrm{rel}}|$) in \textbf{bold};
ties share the bold.
Persistence ignores node features, so its \texttt{noisy\_features\_01} and
\texttt{missing\_features\_20} rows degenerate to the clean numbers (and
$\Delta_{\mathrm{rel}}=0$); EvolveGCN's predictions likewise change only
negligibly there ($|\Delta_{\mathrm{rel}}|<10^{-8}$). Because both models'
inputs are effectively unperturbed under these two feature regimes, we leave
the $\Delta_{\mathrm{rel}}$ column unbolded in these two regimes: a near-zero value
reflects feature-insensitivity, not robustness. In the data-degradation
regimes EvolveGCN attains the smallest $|\Delta_{\mathrm{rel}}|$ yet the worst
HHI and Gini MAE, so its low relative degradation reflects uniformly
weak-but-stable forecasts rather than robustness, as $\Delta_{\mathrm{rel}}$
rewards consistency irrespective of accuracy.}
\label{tab:b6_detail}
\begin{tabular}{l l c c c c c}
\toprule
Regime & Method & PR MAE ($\downarrow$) & HHI MAE ($\downarrow$) & Gini MAE ($\downarrow$) & Rank Corr ($\uparrow$) & $\Delta_{\mathrm{rel}}$ (\textasciitilde 0) \\
\midrule
\multirow{4}{*}{low\_data\_10pct}
& \texttt{m1\_persistence} & 15.94          & 0.0591          & \textbf{0.0023} & \textbf{0.552} & $+0.766$ \\
& \texttt{m3\_evolvegcn}   & 15.23          & 0.0999          & 0.1301          & 0.535          & $\mathbf{+0.276}$ \\
& \texttt{m4\_fm\_only}    & 12.83          & \textbf{0.0586} & 0.0027          & 0.528          & $+0.612$ \\
& \texttt{m5\_fm\_rules}   & \textbf{12.71} & 0.0587          & 0.0027          & 0.528          & $+0.607$ \\
\midrule
\multirow{4}{*}{low\_data\_25pct}
& \texttt{m1\_persistence} & \textbf{9.46}  & 0.0282          & \textbf{0.0020} & \textbf{0.577} & $-0.093$ \\
& \texttt{m3\_evolvegcn}   & 10.22          & 0.0486          & 0.1281          & 0.558          & $\mathbf{-0.027}$ \\
& \texttt{m4\_fm\_only}    & 9.86           & \textbf{0.0279} & 0.0024          & 0.553          & $-0.126$ \\
& \texttt{m5\_fm\_rules}   & 9.95           & \textbf{0.0279} & 0.0024          & 0.553          & $-0.126$ \\
\midrule
\multirow{4}{*}{partial\_graph\_30}
& \texttt{m1\_persistence} & \textbf{9.63}  & \textbf{0.0350} & \textbf{0.0019} & \textbf{0.554} & $+0.068$ \\
& \texttt{m3\_evolvegcn}   & 10.36          & 0.0488          & 0.1236          & 0.537          & $\mathbf{-0.048}$ \\
& \texttt{m4\_fm\_only}    & 12.52          & \textbf{0.0350} & 0.0022          & 0.500          & $+0.082$ \\
& \texttt{m5\_fm\_rules}   & 12.45          & \textbf{0.0350} & 0.0022          & 0.499          & $+0.077$ \\
\midrule
\multirow{4}{*}{noisy\_features\_01}
& \texttt{m1\_persistence} & \textbf{9.77}  & \textbf{0.0322} & \textbf{0.0018} & \textbf{0.584} & $0.000$ \\
& \texttt{m3\_evolvegcn}   & 10.68          & 0.0489          & 0.1326          & 0.563          & $0.000$ \\
& \texttt{m4\_fm\_only}    & 11.38          & 0.0323          & 0.0021          & 0.554          & $-0.005$ \\
& \texttt{m5\_fm\_rules}   & 11.35          & 0.0323          & 0.0021          & 0.554          & $-0.007$ \\
\midrule
\multirow{4}{*}{missing\_features\_20}
& \texttt{m1\_persistence} & \textbf{9.77}  & \textbf{0.0322} & \textbf{0.0018} & \textbf{0.584} & $0.000$ \\
& \texttt{m3\_evolvegcn}   & 10.68          & 0.0489          & 0.1326          & 0.563          & $0.000$ \\
& \texttt{m4\_fm\_only}    & 12.11          & 0.0323          & 0.0023          & 0.559          & $+0.017$ \\
& \texttt{m5\_fm\_rules}   & 12.08          & 0.0323          & 0.0023          & 0.559          & $+0.014$ \\
\bottomrule
\end{tabular}
\end{table*}

At $h{=}4$ under five degradation regimes, EvolveGCN's small relative
degradation reflects uniformly weak-but-stable forecasts (worst HHI and
Gini MAE) rather than genuine robustness, while persistence and the FM
variants remain accurate and are by construction insensitive to the two
feature-perturbation regimes.

\paragraph{Per-budget \texttt{b2\_warning} results (historical event study).}
\label{app:b2-budget}
\begin{table*}[!htbp]
\centering
\footnotesize
\setlength{\tabcolsep}{4pt}
\caption{\texttt{b2\_warning} per-event detail for the shared weighted-degree
monitor (\texttt{h1\_weighted\_degree}) across alert budgets $K\in\{5,10,20\}$.
Recall and F1-warning shown in $10^{-3}$ units (the full DeFi corpus contains
$\sim$4{,}300 protocols, so per-week recall denominators are large). Lead
time is event-level and not budget-sensitive: $5$ weeks for Terra/Luna and
FTX, $4$ weeks for SVB-USDC. \texttt{b2\_warning} sits upstream of the
predictor choice, so a single monitor is shared across all methods
(see \S\ref{sec:results}).}
\label{tab:b2_budget}
\begin{tabular}{l c c c c c c}
\toprule
Event & $K$ & Prec. ($\uparrow$) & Rec.\,$\times 10^{3}$ ($\uparrow$) & F1\textsubscript{w}\,$\times 10^{3}$ ($\uparrow$) & Stab. ($\uparrow$) & $|\mathcal{A}|$ \\
\midrule
\multirow{3}{*}{Terra/Luna}
& 5  & 0.60 & 1.81 & 3.61 & 0.44 & 11 \\
& 10 & 0.70 & 3.26 & 6.49 & 0.62 & 19 \\
& 20 & 0.65 & 7.97 & 15.74 & 0.74 & 36 \\
\midrule
\multirow{3}{*}{FTX}
& 5  & 0.60 & 1.33 & 2.65 & 0.64 & 10 \\
& 10 & 0.70 & 2.39 & 4.77 & 0.72 & 18 \\
& 20 & 0.55 & 4.79 & 9.49 & 0.72 & 39 \\
\midrule
\multirow{3}{*}{SVB-USDC}
& 5  & 1.00 & 1.81 & 3.62 & 0.55 & 10 \\
& 10 & 1.00 & 2.82 & 5.62 & 0.75 & 15 \\
& 20 & 1.00 & 5.64 & 11.22 & 0.81 & 30 \\
\bottomrule
\end{tabular}
\end{table*}

For the three pre-test crisis windows the shared weighted-degree
monitor (\texttt{h1}) achieves median lead times of $4$--$5$ weeks,
with precision $1.000$ at SVB/USDC across all alert budgets
$K\!\in\!\{5,10,20\}$. This is historical monitor evidence on
pre-2025 events, not a test-split result.

\paragraph{Per-regime A1 (data-health gate) isolation.}
\label{app:a1-isolated}
\begin{table*}[!htbp]
\centering
\footnotesize
\setlength{\tabcolsep}{4pt}
\caption{Per-regime data-health gate ablation isolating A1 from A2.
Each row holds $\tau_{\mathrm{conf}}{=}0$ and sweeps $\tau_{\mathrm{data}}$.
``\textit{strict}''~=~$\tau_{\mathrm{data}}{=}0.85$, ``\textit{on}''~=~$0.70$,
``\textit{off}''~=~$0.00$. The four regimes apply targeted edge or feature
degradation on the clean 2025 test split (\texttt{n\_weeks}\,=\,29):
\texttt{extreme} masks both topology and features at 90\%, \texttt{topo}
removes 90\% of edges only, \texttt{feat} corrupts 90\% of node features
only, \texttt{severe} masks both at 80\%. Numbers expand Panel~B of
Table~\ref{tab:ablation}: under degradation the gate-off setting produces
FIR up to $0.541$, while \textit{on}/\textit{strict} can suppress
intervention tickets via safe-mode and drive FIR to $0$.}
\label{tab:a1_isolated}
\begin{tabular}{l l c c c c c c}
\toprule
Regime & Gate & $\tau_{\mathrm{data}}$ & $\overline{\mathrm{dh}}$ & Safe-mode (\%) & Ticket Prec. ($\uparrow$) & FIR ($\downarrow$) & Interventions \\
\midrule
\multirow{3}{*}{extreme}
& strict & 0.85 & 0.587 & 100 & 0.510 & \textbf{0.000} & 0 \\
& on     & 0.70 & 0.587 & 100 & 0.545 & \textbf{0.000} & 0 \\
& off    & 0.00 & 0.587 & 0   & 0.490 & 0.510          & 57 \\
\midrule
\multirow{3}{*}{extreme\_topo}
& strict & 0.85 & 0.637 & 100 & 0.517 & \textbf{0.000} & 0 \\
& on     & 0.70 & 0.637 & 100 & 0.462 & \textbf{0.000} & 0 \\
& off    & 0.00 & 0.637 & 0   & 0.559 & 0.441          & 58 \\
\midrule
\multirow{3}{*}{extreme\_feat}
& strict & 0.85 & 0.755 & 100 & 0.503 & \textbf{0.000} & 0 \\
& on     & 0.70 & 0.755 & 0   & 0.510 & 0.267          & 5 \\
& off    & 0.00 & 0.755 & 0   & 0.503 & 0.267          & 5 \\
\midrule
\multirow{3}{*}{severe}
& strict & 0.85 & 0.751 & 100 & 0.490 & \textbf{0.000} & 0 \\
& on     & 0.70 & 0.751 & 0   & 0.538 & 0.400          & 25 \\
& off    & 0.00 & 0.751 & 0   & 0.538 & 0.541          & 24 \\
\bottomrule
\end{tabular}
\end{table*}

Isolating the data-health gate from the confidence gate
($\tau_{\mathrm{conf}}{=}0$ throughout), the gate-off setting lets FIR climb
to $0.541$ under degradation, while the \textit{on}/\textit{strict} settings
suppress intervention tickets via safe-mode and drive FIR to $0$.

\paragraph{Stress-condition ablation Panels B and C.}
\label{app:ablation-stress}
\begin{table*}[t]
\centering
\footnotesize
\setlength{\tabcolsep}{4pt}
\caption{Component-level ablation of \sysname{} on
\texttt{b5\_decision}. Panel~A shows clean 2025 test-split ablations against
the \texttt{m5\_fm\_rules} baseline; Panels~B--C report targeted stress runs
for mechanisms that are dormant on clean data. Rows follow pipeline order
(Layer~2 evidence then Layer~4 gates). A3 (scenario engine) and A2
(confidence gating) are load-bearing in the clean setting, while A6
(multi-horizon monitoring) and A1 (data-health gating) are safety-reserve
mechanisms whose effects appear under crisis-window dynamics and data
degradation, respectively.}
\label{tab:ablation}
\begin{tabularx}{\textwidth}{@{}l l >{\raggedright\arraybackslash}X c c >{\raggedright\arraybackslash}X@{}}
\toprule
Panel & Ablation & Configuration & Ticket Prec. & FIR & Notes \\
\midrule
\multirow{5}{*}{A: clean test}
 & ---  & full \texttt{m5\_fm\_rules}                                       & 0.600 & 0.000 & baseline \\
 & A6   & single-horizon forecasting ($h{=}4$)                              & 0.600 & 0.000 & no effect (clean data) \\
 & A3   & skip scenario engine                                              & 0.000 & 0.000 & target extraction collapses \\
 & A1   & disable data-health gate ($\tau_{\mathrm{data}}{=}0$)             & 0.600 & 0.000 & no effect (clean data) \\
 & A2   & disable confidence gate ($\tau_{\mathrm{conf}}{=}0$)              & 0.600 & 0.429 & FIR shoots up \\
\midrule
\multirow{2}{*}{B: degraded data}
 & A1 off & 80--98\% feature/edge mask, $\tau_{\mathrm{conf}}{=}0$          & ---   & 0.27--0.60 & false interventions emerge \\
 & A1 on  & same degradation, safe-mode triggered                           & ---   & 0.000      & intervention suppression \\
\midrule
\multirow{2}{*}{C: crisis period}
 & A6: single & horizons $=\{4\}$, crisis windows                         & 0.533--0.624 & 0.000 & alerts: 19--23 \\
 & A6: multi  & horizons $=\{1,4,8,12\}$, same windows                    & 0.533--0.624 & 0.000 & alerts: 73--98 ($\approx4\times$) \\
\bottomrule
\end{tabularx}
\end{table*}

Panel~B: under severe feature/edge masking, FIR ranges over
$[0.27,0.60]$ when the data-health gate is disabled
($\tau_{\mathrm{data}}\!=\!0$; the confidence gate is held off at
$\tau_{\mathrm{conf}}\!=\!0$ throughout Panel~B to isolate A1), and
collapses to $0.000$ when the gate is triggered, by suppressing
intervention tickets.
Panel~C: multi-horizon monitoring ($h\!\in\!\{1,4,8,12\}$) generates
73--98 alerts in the three crisis windows vs 19--23 for
single-horizon ($h\!=\!4$), at unchanged precision range
$0.533$--$0.624$ and zero false interventions in both cases.

\subsection{Qualitative ticket examples}
\label{app:qualitative}
We give two verbatim decision tickets (one representative consistency
run each) behind the aggregate \texttt{b5\_decision} scores. In both
weeks the FM-grounded LLM (\texttt{m7}) escalates while the raw-snapshot
baseline (\texttt{m2}) does not; they differ in whether the escalation
was correct.

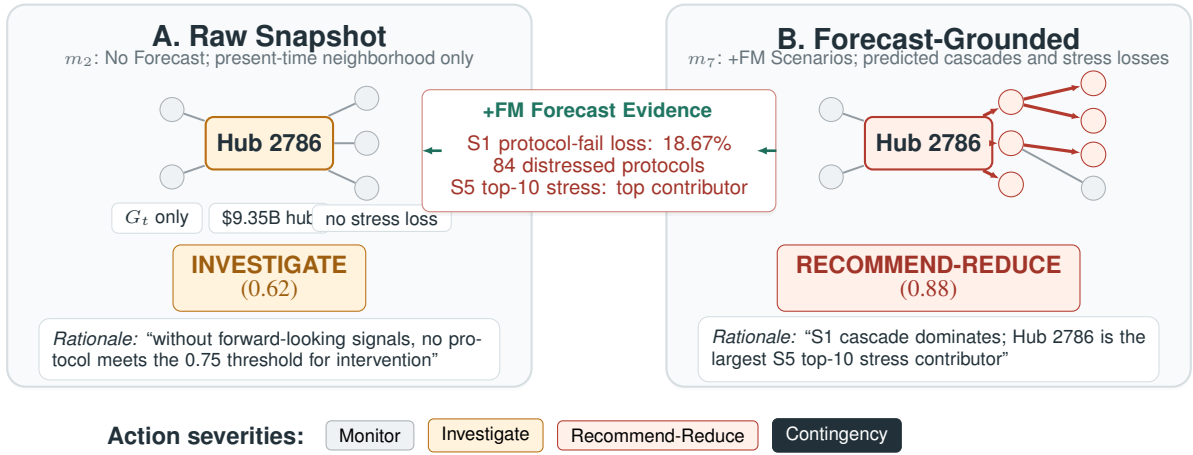
\begin{figure*}[t!]
    \centering
    \definecolor{caseInk}{HTML}{223139}
    \definecolor{caseMuted}{HTML}{65727A}
    \definecolor{caseRule}{HTML}{D6DEE2}
    \definecolor{casePanel}{HTML}{F8FAFB}
    \definecolor{caseGray}{HTML}{EEF2F4}
    \definecolor{caseOrange}{HTML}{B96F13}
    \definecolor{caseOrangeBg}{HTML}{FFF3DE}
    \definecolor{caseRed}{HTML}{B33A2E}
    \definecolor{caseRedBg}{HTML}{FFF0EA}
    \definecolor{caseGreen}{HTML}{248A72}
    \resizebox{0.98\textwidth}{!}{%
    \begin{tikzpicture}[
        font=\sffamily\small,
        panel/.style={
            draw=caseRule, fill=casePanel, rounded corners=7pt,
            line width=0.8pt, minimum width=7.0cm, minimum height=5.1cm,
            drop shadow={opacity=0.035, shadow xshift=0pt, shadow yshift=-1.5pt}
        },
        title/.style={font=\sffamily\bfseries\normalsize, text=caseInk},
        subtitle/.style={font=\sffamily\scriptsize, text=caseMuted},
        chip/.style={
            draw=caseRule, fill=white, rounded corners=2.5pt,
            inner xsep=5pt, inner ysep=2.5pt, font=\sffamily\scriptsize,
            align=center, text=caseInk
        },
        metric/.style={
            draw=caseRed!65, fill=caseRedBg, rounded corners=2.5pt,
            inner xsep=5pt, inner ysep=2.5pt, font=\sffamily\bfseries\scriptsize,
            align=center, text=caseRed!90!black
        },
        evidencebox/.style={
            draw=caseRed!65, fill=caseRedBg, rounded corners=3pt,
            text width=4.3cm, inner xsep=6pt, inner ysep=5pt,
            font=\sffamily\fontsize{7.6pt}{8.4pt}\selectfont,
            align=center, text=caseRed!90!black,
            drop shadow={opacity=0.08, shadow xshift=0pt, shadow yshift=-1.5pt}
        },
        decision/.style={
            draw, rounded corners=3pt, inner xsep=7pt, inner ysep=4pt,
            font=\sffamily\bfseries\footnotesize, align=center
        },
        quote/.style={
            draw=caseRule, fill=white, rounded corners=4pt,
            text width=5.8cm, inner sep=5pt, font=\sffamily\scriptsize,
            align=left, text=caseInk
        },
        hub/.style={rectangle, rounded corners=3pt, draw, thick, minimum width=1.7cm, minimum height=0.7cm, align=center,
            font=\sffamily\bfseries\small, text=caseInk},
        neigh/.style={circle, draw=caseMuted!55, fill=caseGray, minimum size=0.32cm, inner sep=0pt},
        neighalert/.style={circle, draw=caseRed, fill=caseRedBg, minimum size=0.33cm, inner sep=0pt},
        edge norm/.style={draw=caseMuted!70, line width=0.75pt},
        edge alert/.style={draw=caseRed, line width=1.2pt, -{Latex[length=3.4pt,width=2.6pt]}},
        addarrow/.style={draw=caseGreen!80!black, line width=1.0pt, -{Latex[length=4pt,width=3pt]}},
        legbox/.style={draw, rounded corners=2pt, inner xsep=5pt, inner ysep=3pt,
            font=\sffamily\scriptsize}
    ]

    \def\leftX{-4.4}
    \def\rightX{4.4}
    \def\panelY{0.3}

    \node[panel] at (\leftX,\panelY) {};
    \node[panel] at (\rightX,\panelY) {};

    \node[title] at (\leftX,2.40) {A. Raw Snapshot};
    \node[subtitle] at (\leftX,2.12) {$m_2$: No Forecast; present-time neighborhood only};
    \node[title] at (\rightX,2.40) {B. Forecast-Grounded};
    \node[subtitle] at (\rightX,2.12) {$m_7$: +FM Scenarios; predicted cascades and stress losses};

    \begin{scope}[shift={(\leftX,1.0)}]
        \node[hub, fill=caseOrangeBg, draw=caseOrange] (A_hub) at (0,0) {Hub 2786};
        \node[neigh] (A_n1) at (-1.3, 0.45) {};
        \node[neigh] (A_n2) at (-1.3, -0.45) {};
        \node[neigh] (A_n3) at (1.3, 0.6) {};
        \node[neigh] (A_n4) at (1.3, 0) {};
        \node[neigh] (A_n5) at (1.3, -0.6) {};
        \foreach \i in {1,2} \draw[edge norm] (A_n\i) -- (A_hub);
        \foreach \i in {3,4,5} \draw[edge norm] (A_hub) -- (A_n\i);
    \end{scope}

    \node[chip] at (\leftX-1.5,0.0) {$G_t$ only};
    \node[chip] at (\leftX,     0.0) {\$9.35B hub};
    \node[chip] at (\leftX+1.5,0.0) {no stress loss};

    \begin{scope}[shift={(\rightX,1.0)}]
        \node[hub, fill=caseRedBg, draw=caseRed] (B_hub) at (0,0) {Hub 2786};

        \node[neigh] (B_n4) at (-1.3, 0.45) {};
        \node[neigh] (B_n5) at (-1.3, -0.45) {};

        \node[neighalert] (B_n1) at (1.1, 0.55) {};
        \node[neighalert] (B_n2) at (1.1, 0) {};
        \node[neighalert] (B_n3) at (1.1, -0.55) {};

        \node[neighalert] (B_n11) at (2.2, 0.8) {};
        \node[neighalert] (B_n12) at (2.2, 0.3) {};
        \node[neighalert] (B_n21) at (2.2, -0.15) {};
        \node[neigh]      (B_n22) at (2.2, -0.6) {}; 

        \draw[edge norm] (B_n4) -- (B_hub);
        \draw[edge norm] (B_n5) -- (B_hub);
        \draw[edge norm] (B_n2) -- (B_n22);

        \draw[edge alert] (B_hub) -- (B_n1);
        \draw[edge alert] (B_hub) -- (B_n2);
        \draw[edge alert] (B_hub) -- (B_n3);
        \draw[edge alert] (B_n1) -- (B_n11);
        \draw[edge alert] (B_n1) -- (B_n12);
        \draw[edge alert] (B_n2) -- (B_n21);
    \end{scope}

    \node[evidencebox, fill=white] (FMbox) at (0, 0.9) {
        \textcolor{caseGreen!85!black}{\textbf{+FM Forecast Evidence}}\\[4pt]
        S1 protocol-fail loss: 18.67\%\\
        84 distressed protocols\\
        S5 top-10 stress: top contributor
    };
    \draw[addarrow] (-2.1, 0.9) -- (FMbox.west);
    \draw[addarrow] (FMbox.east) -- (2.1, 0.9);

    \node[decision, fill=caseOrangeBg, draw=caseOrange, text=caseOrange!85!black]
        at (\leftX,-0.8) {INVESTIGATE\\[-1pt]\normalfont (0.62)};
    
    \node[decision, fill=caseRedBg, draw=caseRed, text=caseRed!90!black]
        at (\rightX,-0.8) {RECOMMEND-REDUCE\\[-1pt]\normalfont (0.88)};

    \node[quote] at (\leftX,-1.75) {
        \textit{Rationale:} ``without forward-looking signals, no protocol meets the
        0.75 threshold for intervention''
    };
    \node[quote] at (\rightX,-1.75) {
        \textit{Rationale:} ``S1 cascade dominates; Hub 2786 is the largest
        S5 top-10 stress contributor''
    };

    \begin{scope}[shift={(-3.8,-2.9)}]
        \node[font=\sffamily\bfseries\footnotesize, anchor=east, text=caseInk] at (0,0) {Action severities:};
        \node[legbox, fill=caseGray, draw=caseMuted!55, anchor=west] (L_mon) at (0.15,0) {Monitor};
        \node[legbox, fill=caseOrangeBg, draw=caseOrange, anchor=west] (L_inv) at ([xshift=0.18cm]L_mon.east) {Investigate};
        \node[legbox, fill=caseRedBg, draw=caseRed, anchor=west] (L_rec) at ([xshift=0.18cm]L_inv.east) {Recommend-Reduce};
        \node[legbox, fill=caseInk, draw=caseInk, text=white, anchor=west] at ([xshift=0.18cm]L_rec.east) {Contingency};
    \end{scope}

    \end{tikzpicture}%
    }
    \caption{Hub 2786 case (week 2025-03-24). The expanded evidence names S1 protocol failure and S5 top-10 stress; forecast grounding moves the action from \textsc{Investigate} to \textsc{Recommend-Reduce}.}
    \label{fig:qualitative}
\end{figure*}

\paragraph{Success: 2025-03-24 (FIR $0$, target precision $0.83$, Opus-4.8 judge \texttt{m7}$=4$, \texttt{m2}$=2$).}
On the same hub (node~2786, $9.35$B exposure), \texttt{m2} stays at
\emph{moderate} / \textsc{Investigate} ($0.62$) because, in its own
words, ``without forward-looking signals no protocol meets the $0.75$
threshold for intervention.'' \texttt{m7} reads the same hub as
\emph{elevated} / \textsc{Recommend-Reduce} ($0.88$): ``largest predicted
exposure ($9.35$B), dominant in S1 (single-failure cascade causes
$18.67\%$ system loss, $84$ distressed, $282$ affected) and largest
contributor to S5 correlated stress.'' All three reduced hubs
($2786$, $2269$, $118$) fall inside the realized top-$5\%$ stress set.

\paragraph{Over-intervention: 2025-04-28 (FIR $1.0$, target precision $0$, Opus-4.8 judge \texttt{m7}$=3$).}
Here \texttt{m7} escalates the three largest hubs ($2269$/$2786$/$1599$)
to \textsc{Recommend-Reduce}, citing an S1 top-protocol failure of
$22.14\%$ system loss. The grounding is real but the targets are wrong:
none of the named hubs is in the realized top-$5\%$ stress set, which
that week was dominated by long-tail protocols. The Opus-4.8 judge's own
rationale names the failure mode: the report ``cites specific network
metrics \dots\ and provides plausible cascade scenario reasoning, but
none of its named targets match the ground-truth top-$5\%$ systemic risk
set, indicating poor predictive accuracy despite well-grounded data
references.'' Across the test split this decoupling between explanation
quality and target correctness is only weak (Spearman between Opus-4.8
quality and $1-\mathrm{FIR}$ is $0.31$), but it is the mechanism behind
\texttt{m7}'s aggregate FIR of $0.437$, which persists across decision
models (Table~\ref{tab:stats_ci}).

\subsection{Judge panel}
\label{app:judge-panel}

The cross-decision-LLM panel
(Table~\ref{tab:stats_ci}) varies the decision model; here we report the
cross-family judge panel that fixes the decisions and varies the judge.

\begin{table}[t]
\centering\small
\setlength{\tabcolsep}{4pt}
\caption{Cross-family judge panel on the same Opus~4.7 decisions
($n\!=\!29$). Gemini and GPT use $\mathrm{reasoning}\!=\!
\{\mathrm{effort:minimal}\}$ so the thinking budget matches the
Anthropic baseline. \textsuperscript{\dag}~Haiku~4.5 is a
tier-below weak-judge reference.}
\label{tab:judge-panel}
\begin{tabular}{@{}lcccc@{}}
\toprule
Method & Opus~4.8 & Gemini~2.5~Pro & GPT-5.5 & Haiku~4.5\textsuperscript{\dag} \\
\midrule
\texttt{m2} & 2.24 & 2.69 & 1.90 & 2.03 \\
\texttt{m6} & 2.41 & 2.55 & 2.45 & 2.48 \\
\texttt{m7} & \textbf{2.45} & \textbf{2.90} & 2.45 & 2.69 \\
\bottomrule
\end{tabular}
\end{table}

\paragraph{Complete headline metrics.}
Table~\ref{tab:headline-full} reports the full \texttt{b5\_decision}
comparison, restoring the grounding and target-stability columns dropped
from the compact body panel (Table~\ref{tab:rq2}). Grounding is
$1.000$ for every LLM-bearing system, so the over-intervention captured by
FIR is not a citation-fabrication artifact. Target stability (Jaccard
overlap of ticket targets and severities across three temperature-$0$
calls) is lowest for the FM+rules stack (\texttt{m5}, $0.257$); the
FM-grounded LLM variants (\texttt{m6}/\texttt{m7}) stay within the range of
the snapshot baselines.

\begin{table*}[!htbp]
\centering\small
\setlength{\tabcolsep}{3pt}
\caption{Complete \texttt{b5\_decision} comparison on the frozen 2025 test
split ($n\!=\!29$); the full version of Table~\ref{tab:rq2}. Grnd.\ is
grounding (judge-confirmed evidence citation) and Stab.\ is target
stability. Dashes are undefined: \texttt{m1}/\texttt{m5} emit no
LLM-judged tickets, and \texttt{m1}/\texttt{m2}/\texttt{m5} emit no
intervention-level ticket that could misfire.}
\label{tab:headline-full}
\begin{tabular}{@{}lllccccccc@{}}
\toprule
ID & Pred. & Decision & Prec.$\uparrow$ & Rec.$\uparrow$ & F1$\uparrow$ & Judge$\uparrow$ & FIR$\downarrow$ & Grnd.$\uparrow$ & Stab.$\uparrow$ \\
\midrule
\texttt{m1} & persist.   & rules               & \textbf{0.720} & 0.004 & 0.0076          & ---            & ---            & ---   & 0.514 \\
\texttt{m2} & ---        & LLM (no FM)         & 0.575          & 0.009 & 0.0184          & 2.24           & ---            & 1.000 & 0.532 \\
\texttt{m5} & \fmname{}  & rules               & 0.600          & 0.010 & 0.0190          & ---            & ---            & ---   & 0.257 \\
\texttt{m6} & \fmname{}  & LLM                 & 0.570          & 0.012 & \textbf{0.0241} & 2.41           & 0.448          & 1.000 & 0.488 \\
\texttt{m7} & \fmname{}  & LLM + safety gate   & 0.580          & 0.012 & 0.0234          & \textbf{2.45}  & 0.437          & 1.000 & 0.435 \\
\bottomrule
\end{tabular}
\end{table*}

\subsection{Statistical significance and matched-budget analysis}
\label{app:stats}

\begin{table*}[t]
\centering
\footnotesize
\setlength{\tabcolsep}{5pt}
\caption{95\% bootstrap CIs for the decision-quality headline
numbers (decision LLM in parentheses; judge is Claude~Opus~4.8
throughout). \textsuperscript{\ddag}\texttt{m1}'s per-week
distribution comes from a local re-run of the released harness:
its aggregate F1 ($0.0081$) brackets the published GPU-run value
($0.0076$), and the recall@$k$ and stressed-pool size reproduce
exactly, but with ${<}1$ ticket per week a few borderline tickets
flip across environments (re-run precision $0.633$ vs published
$0.720$), so the \texttt{m1} interval should be read as
order-of-magnitude.}
\label{tab:stats_ci}
\begin{tabular}{@{}lccc@{}}
\toprule
Configuration & F1 [95\% CI] & FIR [95\% CI] & Judge [95\% CI] \\
\midrule
\texttt{m1} rules\textsuperscript{\ddag} & 0.0081 [0.0041, 0.0123] & 0.000 & --- \\
\texttt{m2} (Opus~4.7)       & 0.0184 [0.0142, 0.0223] & 0.000 & 2.24 [2.10, 2.41] \\
\texttt{m6} (Opus~4.7)       & 0.0241 [0.0201, 0.0281] & 0.448 [0.293, 0.601] & 2.41 [2.21, 2.62] \\
\texttt{m7} (Opus~4.7)       & 0.0234 [0.0194, 0.0271] & 0.437 [0.282, 0.592] & 2.45 [2.24, 2.66] \\
\texttt{m6} (Sonnet~4.6)     & 0.0276 [0.0225, 0.0324] & 0.379 [0.236, 0.534] & 2.52 [2.28, 2.79] \\
\texttt{m7} (Sonnet~4.6)     & 0.0288 [0.0236, 0.0335] & 0.374 [0.224, 0.529] & 2.66 [2.41, 2.90] \\
\texttt{m6} (Gemini~2.5~Pro) & 0.0129 [0.0103, 0.0155] & 0.155 [0.034, 0.293] & 2.21 \\
\texttt{m7} (Gemini~2.5~Pro) & 0.0139 [0.0111, 0.0167] & 0.190 [0.069, 0.328] & 2.38 \\
\bottomrule
\end{tabular}
\end{table*}

All headline metrics are means over the $n\!=\!29$ scored test weeks. We
report 95\% percentile bootstrap confidence intervals ($10{,}000$ resamples
of weeks with replacement; F1 recomputed from the resampled mean precision
and recall) and two-sided paired sign-flip permutation tests ($20{,}000$
permutations, methods aligned by week). Analysis code is released with the
harness (\texttt{scripts/bootstrap\_stats.py},
\texttt{scripts/matched\_budget.py}).

\begin{table}[t]
\centering
\footnotesize
\setlength{\tabcolsep}{5pt}
\caption{Two-sided paired permutation $p$-values for the key
comparisons. Significant effects ($p<0.05$) in \textbf{bold}.}
\label{tab:stats_perm}
\begin{tabular}{@{}llr@{}}
\toprule
Comparison & Metric & $p$ \\
\midrule
\texttt{m6} vs \texttt{m2} (FM signal)        & F1   & \textbf{0.0002} \\
\texttt{m7} vs \texttt{m1} (full stack)       & F1   & \textbf{$<$0.0001} \\
\texttt{m7} vs \texttt{m6} (gate)             & F1   & 0.41 \\
\texttt{m7}$\times$Sonnet vs \texttt{m7}$\times$Opus & F1 & \textbf{0.0004} \\
\texttt{m6} vs \texttt{m2}                    & FIR  & \textbf{$<$0.0001} \\
\texttt{m7}$\times$Sonnet vs \texttt{m7}$\times$Opus & FIR & 0.22 \\
\texttt{m6}$\times$Sonnet vs \texttt{m6}$\times$Opus & FIR & 0.12 \\
\texttt{m6} vs \texttt{m2} (Opus~4.8 judge)   & Judge & 0.23 \\
\texttt{m6} vs \texttt{m2} (GPT-5.5 judge)    & Judge & \textbf{0.0002} \\
\texttt{m6} vs \texttt{m2} (Gemini judge)     & Judge & 0.77 \\
\texttt{m7} vs \texttt{m6} (Opus~4.8 judge)   & Judge & 1.00 \\
\texttt{m7}$\times$Sonnet vs \texttt{m7}$\times$Opus & Judge & 0.21 \\
\bottomrule
\end{tabular}
\end{table}

The load-bearing quantitative claims (the FM signal's F1 lift, the full
stack's F1 lift over rules, the Sonnet~4.6 F1 improvement, and the
over-intervention cost) are significant at $p\!<\!0.001$. The
explanation-quality lift of the FM-grounded variants is judge-dependent
(significant under GPT-5.5, positive but not significant under Opus~4.8 at
$p\!=\!0.23$, absent under the noisiest Gemini judge), so we read the judge
axis as directional evidence. The safety gate's effects on F1 and judge
score are indistinguishable from noise on clean data, consistent with its
characterisation as a reserve mechanism (\S\ref{sec:lessons}).

\paragraph{Matched-budget recall@$k$.}
\label{app:stats-budget}
Methods emit different numbers of ticket targets per week (\texttt{m1}
$2.1$, \texttt{m2} $4.9$, \texttt{m7} $6.3$, \texttt{m6} $6.5$,
\texttt{m7}$\times$Sonnet $7.6$), and unbudgeted recall mechanically
favours methods that flag more protocols. To separate target \emph{quality}
from target \emph{volume}, we truncate every method to its top-$k$ targets
per week (LLM variants ranked by their own emitted risk score; \texttt{m1}
by ticket confidence) and score against the same stressed pool.

\begin{table*}[!tb]
\centering
\footnotesize
\setlength{\tabcolsep}{4pt}
\caption{Matched-budget recall@$k$ ($\times 10^{3}$) and
precision@$k$ on the 2025 test split. Every LLM variant beats the
rules baseline at all $k$ ($p\!\leq\!0.006$ vs \texttt{m1}). The
FM-grounded and raw-snapshot variants are indistinguishable through
the head ($k\!\leq\!5$, all pairwise $p\!\geq\!0.28$); they diverge
only once \texttt{m2} exhausts its targets, where the FM variants
recover more stressed protocols at the same precision ($k\!=\!7$,
$p\!\leq\!0.0003$). \textsuperscript{\dag}saturated, no week reaches
depth $k$, so the cell repeats the method's full-set value.
\texttt{m1} row from the local re-run (Table~\ref{tab:stats_ci} note).}
\label{tab:matched_budget}
\begin{tabular}{@{}lcccccccc@{}}
\toprule
& \multicolumn{2}{c}{$k\!=\!1$} & \multicolumn{2}{c}{$k\!=\!3$} & \multicolumn{2}{c}{$k\!=\!5$} & \multicolumn{2}{c}{$k\!=\!7$} \\
\cmidrule(lr){2-3}\cmidrule(lr){4-5}\cmidrule(lr){6-7}\cmidrule(lr){8-9}
Method & R & P & R & P & R & P & R & P \\
\midrule
\texttt{m1} rules            & 0.71 & 0.24 & 2.45 & 0.26 & 4.05 & 0.26 & 4.05\textsuperscript{\dag} & 0.26 \\
\texttt{m2} (Opus~4.7)       & 1.53 & 0.52 & 5.70 & 0.59 & 9.08 & 0.58 & 9.35\textsuperscript{\dag} & 0.58 \\
\texttt{m6} (Opus~4.7)       & 1.64 & 0.55 & 5.30 & 0.56 & 9.63 & 0.59 & \textbf{12.14} & 0.57 \\
\texttt{m7} (Opus~4.7)       & 1.64 & 0.55 & 5.30 & 0.56 & 9.13 & 0.57 & \textbf{11.95} & 0.58 \\
\texttt{m7} (Sonnet~4.6)     & 1.64 & 0.55 & 5.30 & 0.56 & 9.70 & 0.59 & \textbf{13.26} & 0.58 \\
\bottomrule
\end{tabular}
\end{table*}

The matched-budget view decomposes the headline F1 result. Against the
rules baseline, LLM targeting is genuinely better per target. At every $k$
precision@$k$ roughly doubles ($0.52$--$0.59$ vs $0.24$--$0.28$) and
recall@$k$ is significantly higher ($p\!\leq\!0.006$). Against the
raw-snapshot LLM, the FM's contribution is \emph{not} a higher per-target
hit rate. Through the head the two are statistically indistinguishable
(all pairwise $p\!\geq\!0.28$ at $k\!\leq\!5$), and \texttt{m2} even leads
slightly at $k\!=\!3$. The divergence appears only past \texttt{m2}'s
target ceiling. The raw-snapshot model exhausts its targets at about five
per week (no week reaches $k\!=\!7$), so its recall@$k$ saturates at
$9.35\times10^{-3}$, whereas the FM variants keep emitting targets and at
$k\!=\!7$ recover significantly more stressed protocols
($12.14$ vs $9.35$, $p\!=\!0.0001$) at undiminished precision. These extra
targets are genuine detections rather than padding: the marginal precision
of the rank-$6$-to-$8$ tail that only the FM variants populate is $0.535$
for \texttt{m6} and $0.622$ for \texttt{m7}, comparable to their head
precision ($\sim\!0.56$). This larger target set at flat precision is the
FM's contribution to the headline recall and F1 lift (Table~\ref{tab:rq2}),
and the expanded tail is also where over-intervention concentrates. It does
so across decision models, escalated to intervention-level severity at a
comparable rate under Opus~4.7 ($\mathrm{FIR}\!=\!0.448$) and Sonnet~4.6
($\mathrm{FIR}\!=\!0.379$).

\section{Prompt Templates}
\label{app:prompt-templates}

The released audit logs contain the fully instantiated prompts and model
outputs for each evaluated week. For reproducibility, we also report the
templates used by the decision and judge calls below. Placeholders in braces
are filled deterministically from the evidence bundle; line wrapping is for
typesetting only. The prompts expose protocol-level aggregate summaries and
derived evidence, not raw transaction logs, wallet-level records, full edge
lists, or node-feature tables.

\noindent\textbf{Shared decision system prompt.}
\par\nobreak\vspace{0.25\baselineskip}
\begin{lstlisting}[style=prompttemplate]
You are a DeFi systemic risk analyst for a regulatory supervisory body.
Your job: given network data, identify protocols at elevated risk of
distress over the next {horizon} weeks and recommend supervisory actions.

You MUST respond with valid JSON only.
Schema:
{
  "risk_level": "low" | "moderate" | "elevated" | "critical",
  "target_protocols": [
    {
      "protocol": "<name>",
      "risk_score": <float 0-1>,
      "action": "Monitor" | "Investigate" |
                "Recommend-Reduce" | "Contingency",
      "reason": "<specific explanation citing input data>"
    }
  ],
  "rationale": "<2-3 sentence assessment citing metrics and numbers>"
}

Rules:
- Only flag protocols you genuinely believe are at elevated risk.
- Cite specific numbers from the provided data.
- Use exact protocol names as they appear in the data.
- Order target_protocols by risk_score descending.
- Recommend-Reduce requires elevated/critical overall risk and
  protocol risk_score >= 0.75.
- Contingency requires critical overall risk and risk_score >= 0.90.
- If the action constraints are not met, use Investigate.
\end{lstlisting}

\noindent\textbf{Forecast-grounded user prompt (\texttt{m6}/\texttt{m7}).}
\par\nobreak\vspace{0.25\baselineskip}
\begin{lstlisting}[style=prompttemplate]
Current DeFi network analysis ({date}), forecast horizon = {horizon} weeks.

== FM MODEL PREDICTIONS (predicted graph G_{t+{horizon}}) ==
Nodes: {n_nodes} protocols, Edges: {n_edges} weighted exposure links

Top-10 protocols by predicted exposure weight:
{top_protocols}

== PREDICTED NETWORK RISK METRICS ==
{metrics}

== STRESS SCENARIO ANALYSIS (applied to predicted graph) ==
{scenarios}

Based on these FM model predictions, metrics, and stress analysis, identify
which protocols are at elevated risk over the next {horizon} weeks.
\end{lstlisting}

\noindent\textbf{Snapshot-only user prompt (\texttt{m2}).}
\par\nobreak\vspace{0.25\baselineskip}
\begin{lstlisting}[style=prompttemplate]
Current DeFi network state ({date}), forecast horizon = {horizon} weeks.

NOTE: You do NOT have access to any predictive model. You must reason from
the current network snapshot only. There are no forward-looking predictions.

== CURRENT NETWORK SNAPSHOT ==
Protocols: {n_nodes} total
Edges: {n_edges} weighted exposure links
Total network weight: {total_weight}

Top-10 protocols by current exposure weight:
{top_protocols}

Category breakdown:
{category_summary}

== CURRENT NETWORK METRICS ==
{metrics}

Based on this current network state, identify which protocols are at
elevated risk over the next {horizon} weeks.
\end{lstlisting}

\noindent\textbf{LLM-as-judge prompt.}
\par\noindent
The judge prompt uses only aggregate evaluation context: week, horizon,
network-scale counts, aggregate metric summaries, the top-loss ground-truth
labels used for scoring, and the report being judged. The fields
\texttt{n\_nodes} and \texttt{n\_edges} are graph-size counts, not node or edge
records.
\par\nobreak\vspace{0.25\baselineskip}
\begin{lstlisting}[style=prompttemplate]
SYSTEM:
You are an expert evaluator of DeFi risk analysis reports. You will compare
two risk assessments for the same week and rate the quality of the SECOND
one (Report B) on a scale of 1-5.

Respond with JSON only:
{
  "quality_score": <int 1-5>,
  "reasoning": "<1-2 sentence justification>"
}

USER:
Week: {date}, horizon: {horizon} weeks

== INPUT DATA SUMMARY ==
Network: {n_nodes} nodes, {n_edges} edges
Key metrics: {metrics_summary}

== GROUND TRUTH ==
{n_stressed} protocols are in the top-5% set ranked by absolute weight loss
over the horizon: {stressed_list}

== REPORT TO EVALUATE ==
Risk level: {risk_level}
Targets: {llm_targets}
Rationale: {rationale}

Rate the report's quality (1-5).
\end{lstlisting}

\end{document}